\documentclass[runningheads,a4paper]{llncs}

\usepackage{amssymb}
\usepackage{amsmath}
\usepackage{graphicx}
\usepackage{subcaption}
\usepackage{color}
\usepackage{url}
\urldef{\mailsa}\path|{alfred.hofmann, ursula.barth, ingrid.haas, frank.holzwarth,|
\urldef{\mailsb}\path|anna.kramer, leonie.kunz, christine.reiss, nicole.sator,|
\urldef{\mailsc}\path|erika.siebert-cole, peter.strasser, lncs}@springer.com|    
\newcommand{\keywords}[1]{\par\addvspace\baselineskip
\noindent\keywordname\enspace\ignorespaces#1}

\def\D{\mathrm{d}}

\begin{document}

\mainmatter  

\title{Noisy Softplus: an activation function that enables SNNs to be trained as ANNs}

\titlerunning{ANN Trained SNN by Noisy Softplus}

%
%
\author{Qian Liu$^{1}$%
\and Yunhua Chen$^{2,*}$\and Steve Furber$^{1}$}
%

\institute{
	$^{1}$University of Manchester, Manchester, UK.\\
	\{qian.liu-3, steve.furbe\}@manchester.ac.uk\\
	$^{2}$Guangdong University of Technology, Guangzhou, China.\\
	yhchen@gdut.edu.cn
	}


%
%

\toctitle{Lecture Notes in Computer Science}
\tocauthor{Authors' Instructions}
\maketitle

\begin{abstract}
 We extended the work of proposed activation function, Noisy Softplus~\cite{liu2016noisy}, to fit into training of layered up spiking neural networks~(SNNs).
 Thus, any ANN employing Noisy Softplus neurons, even of deep architecture, can be trained simply by the traditional algorithm, for example Back Propagation~(BP), and the trained weights can be directly used in the spiking version of the same network without any conversion.
 Furthermore, the training method can be generalised to other activation units, for instance Rectified Linear Units (ReLU), to train deep SNNs off-line.
 This research is crucial to provide an effective approach for SNN training, and to increase the classification accuracy of SNNs with biological characteristics and to close the gap between the performance of SNNs and ANNs.
\end{abstract}

\section{Introduction}
DNNs are the most promising research field in computer vision, even exceeding human-level performance on image classification tasks~\cite{he2015delving}.
To investigate whether brains might work similarly on vision tasks, these powerful DNN models have been converted to SNNs.
In addition, spiking DNNs offer the prospect of neuromorphic systems that combine remarkable performance with energy-efficient training and operation.

Theoretical studies have shown that biologically-plausible learning, e.g. Spike-Timing-Dependent Plasticity (STDP), could approximate a stochastic version of powerful machine learning algorithms
such as 
Contrastive Divergence~\cite{neftci2013event}, Markov Chain Monte Carlo~\cite{buesing2011neural} and Gradient Descent~\cite{o2016deep}.
It is the stochasticity, in contrast to the continuously differentiable functions used by ANNs, that is intrinsic to the event-based spiking process, making network training difficult.
In practice, ANNs use neuron and synapse models very different from biological neurons, and it remains an unsolved problem to develop SNNs with equivalent performance.


Conversely, the offline training of an ANN, which is then mapped to an SNN, has shown near loss-less conversion and state-of-the-art classification accuracy.
This research aims to prove that SNNs are equally capable as their non-spiking rivals of pattern recognition, and at the same time are more biologically realistic and energy-efficient.
Jug et al.~\cite{Jug_etal_2012} first proposed the use of the Siegert function to replace the sigmoid activation function in training Restricted Boltzmann Machine (RBM).
The Siegert units map incoming currents driven by Poisson spike trains to the response firing rate of a Leaky Integrate-and-Fire (LIF) neuron.
The ratio of the spiking rate to its maximum is equivalent to the output of a sigmoid neuron.
A spiking Deep Belief Network (DBN)~\cite{Stromatias2015scalable} of four layers of RBMs was implemented on neuromorphic hardware, SpiNNaker~\cite{furber2014spinnaker}, to recognise hand written digits in real time.

However, cortical neurons seldom saturate their firing rate as sigmoid neurons.
Thus ReLU were proposed to replace sigmoid neurons and surpassed the performance of other popular activation units thanks to their advantage of sparsity~\cite{glorot2011deep} and robustness towards the vanishing gradient problem. 
Recent developments on ANN-trained SNN models has focused on using ReLU units and converting trained weights to fit in SNNs.
Better performance~\cite{cao2015spiking,diehl2015fast} than Siegert-trained RBM has been demonstrated in Spiking ConvNets, but this employed simple integrate and fire (IF) neurons without leakage.
The training used only ReLUs and zero bias to avoid negative outputs, and applied a deep learning technique, dropout~\cite{srivastava2014dropout}, to increase the classification accuracy.
Normalising the trained weights for use on an SNN employing IF neurons only was relatively straightforward and maintained classification accuracy.
This work was extended to a Recursive Neural Network (RNN)~\cite{diehl2016conversion} and run on the TrueNorth~\cite{merolla2014million} neuromorphic hardware platform.

Except for the popular, simplified version of ReLU, $max(0,\sum w x)$, the other implementation of $\log(1+e^x)$, ``Softplus'', is more biologically realistic.
Recent work~\cite{hunsberger2015spiking} proposed the Soft LIF response function for training SNNs, which is equivalent to Softplus activation of ANNs.

We have discussed the difficulty of SNN training, and considered existing state-of-the-art solutions.
We then propose an activation function, Noisy Softplus, and demonstrate how it fits the network dynamics composed of spiking neurons in Section~\ref{sec:af_model}.
Section~\ref{sec:ann_train_snn} will illustrate a complete SNN training method which employs Noisy Softplus activations and can be generalised to other activation functions.
To validate the classification accuracy, a convolutional network (ConvNet) is trained on the MNIST database following this training mechanism and tested directly on an SNN in Section~\ref{sec:iconipResult}.

\section{Modelling The Activation Function}
\label{sec:af_model}
The existing work of modelling the response firing activity of LIF neurons using Siegert~\cite{Jug_etal_2012} function has suggested an approach to modelling activation functions for spiking neurons whose input is a synaptic current generated by spike arrivals and whose output is the firing rate of a sequence of spikes.
This section will start from the neural science background to demonstrate those physical quantities used in the Siegert formula.
Then we will compare and show the difference between analytical estimation and practical simulations of spiking neurons.
Consequently, we propose the new activation function, Noisy Softplus, to replace Siegert function for modelling the practical spiking activities of LIF neurons.

\subsection{Neural Science Background}
The LIF neuron model follows the following membrane potential dynamics:
\begin{equation}
\tau_m \frac{\D V}{\D t}=V_{rest} - V + R_{m} I(t) ~.
\label{equ:LIF_V}
\end{equation}
The membrane potential $V$ changes in response to the input current $I$, starting at the resting membrane potential $V_{rest}$, where the membrane time constant is $\tau_m = R_mC_m$, $R_m$ is the membrane resistance and $C_m$ is the membrane capacitance.
The central idea in converting spiking neurons to activation units lies in the response function of a neuron model.
Given a constant current injection $I$, the response function, i.e. firing rate, of the LIF neuron is:
\begin{equation}
\lambda_\mathit{out}=
\left [ t_\mathit{ref}-\tau_m\log \left ( 1-\frac{V_{th}-V_\mathit{rest}}{IR_m}  \right )\right ]^{-1}, \textrm{~when~} IR_m>V_{th}-V_{rest},
\label{equ:consI}
\end{equation}
otherwise the membrane potential cannot reach the threshold $V_{th}$ and the output firing rate is zero. 
The absolute refractory period $t_\mathit{ref}$ is included, during which period synaptic inputs are ignored.

However, in practice, a noisy current generated by the random arrival of spikes, rather than a constant current, flows into the neurons.
The noisy current is typically treated as a sum of a deterministic constant term, $I_{const}$, and a white noise term, $I_{noise}$.
Thus the value of the current is Gaussian distributed with $m_I$ mean and ${s_I}^2$ variance.
The white noise is a stochastic process $\xi(t)$ with mean 0 and variance 1, which is delta-correlated, i.e., the process is uncorrelated in time so that a value $\xi(t)$ at time $t$ is totally independent on the value at any other time $t'$.
Therefore, the noisy current can be seen as:
\begin{equation}
I(t) = I_{const}(t)+I_{noise}(t) = m_I + s_I\xi(t)~~,
\label{equ:noisyI}
\end{equation}
and accordingly, Equation~(\ref{equ:LIF_V}) becomes:
\begin{equation}
\frac{\D V}{\D t}=\frac{V_{rest} - V}{\tau_m } + \frac{m_I}{C_m} + \frac{s_I\xi(t)}{C_m}~~.
\label{equ:LIF_V2}
\end{equation}

We then multiply the both sides of Equation~(\ref{equ:LIF_V2}) by a short time step $\D t$, the stochastic differential equation of the membrane potential satisfies an Ornstein-Uhlenbeck process:
\begin{equation}
\begin{aligned}
\D V&= \frac{V_{rest} - V}{\tau_m }\D t + \frac{m_I}{C_m} \D t + \frac{s_I}{C_m}  \D W_t \\
&=\frac{V_{rest} - V}{\tau_m }\D t + \frac{m_I}{C_m} \D t + \frac{s_I \sqrt{\D t}}{C_m} \xi(t)  \\
&=\frac{V_{rest} - V}{\tau_m }\D t + \mu \D t + \sigma \xi(t) ~~. 
\end{aligned}
\end{equation}	
The last term $\D W_t$ is a Wiener process, that $W_{t +\D t} - W_{t}$ obeys Gaussian distribution with mean 0 and variance $\D t$.
The instantaneous mean $\mu$ and variance $\sigma^2$ of the change in membrane potential characterise the statistics of $V$ in a short time range, and they can be derived from the statistics of the noisy current:
\begin{equation}
\mu =\dfrac{m_I}{C_m}, ~~~~~ \sigma = \dfrac{s_I \sqrt{\D t}}{C_m}~.
\end{equation}
The response function~\cite{rauch2003neocortical,la2008response} of the LIF neuron to a noisy current, also known as Siegert formula~\cite{siegert1951first}, is driven by the $\mu$ and $\sigma$:
\begin{equation}
\lambda_\mathit{out}=
\left [ t_\mathit{ref}+\tau_m \int_{\frac{V_\mathit{rest}-\mu \tau_m }{\sigma \sqrt{\tau_m}}}^{\frac{V_{th}-\mu \tau_m }{\sigma \sqrt{\tau_m}}} \sqrt{\pi} \exp(u^{2}) (1+erf(u)) \D u \right ]^{-1} ~,
\label{equ:siegert}
\end{equation}

\begin{figure}[bt]
	\centering
	\includegraphics[width=0.7\textwidth]{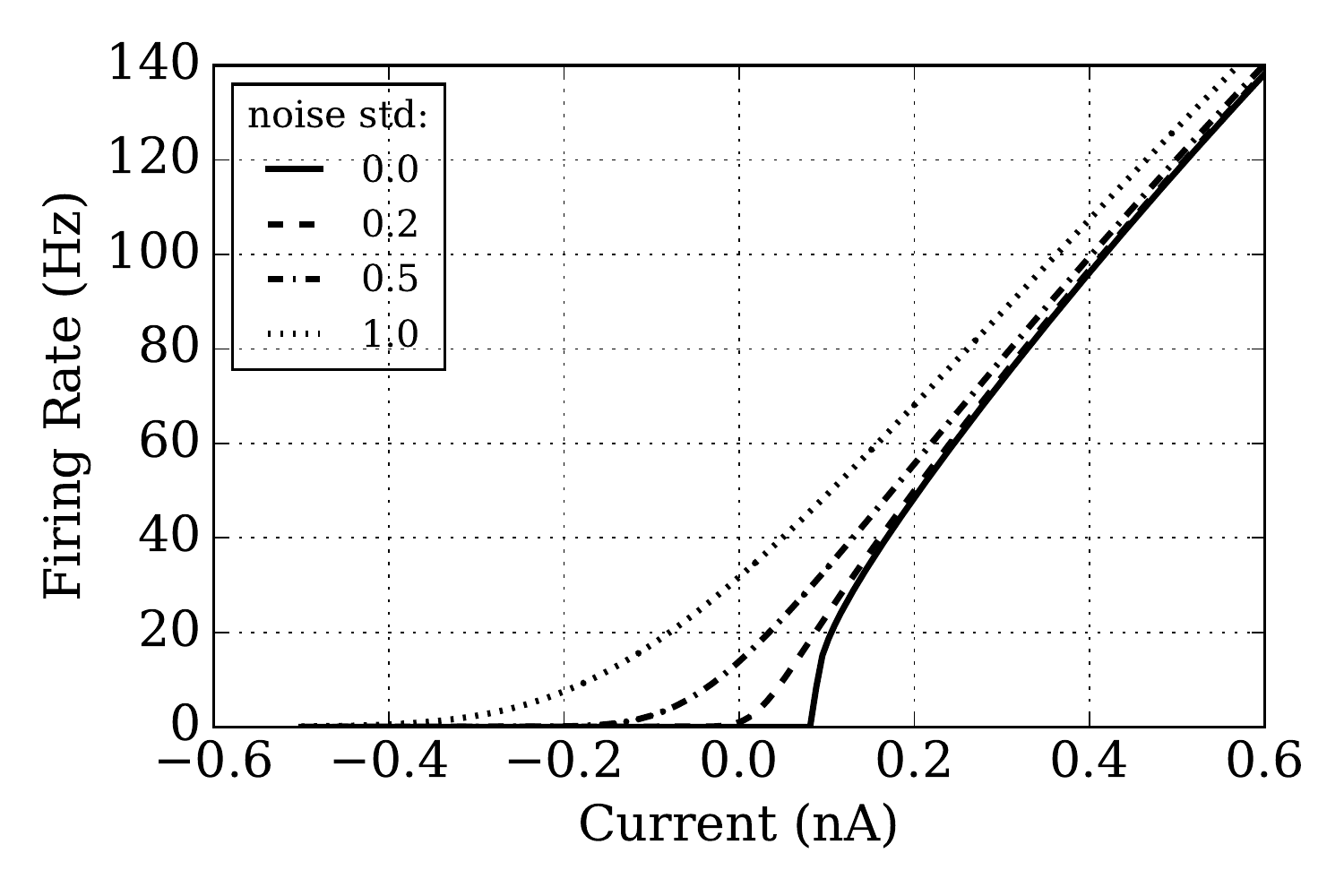}
	\caption{Response function of the LIF neuron with noisy input currents with different standard deviations.}
	\label{Fig:physics}
\end{figure}

Figure~\ref{Fig:physics} shows the response curves (Equation~(\ref{equ:siegert})) of a LIF neuron driven by noisy currents with Gaussian noise of $m_I$ mean and $s_I$ standard deviation.
The parameters of the LIF neuron are all biologically valid (see the listed values in Table~\ref{tbl:pynnConfig}), and the same parameters are used throughout this paper.
The solid (zero noise) line in Figure~\ref{Fig:physics} illustrates the response function of such an LIF neuron injected with constant current, which inspired the proposal of ReLUs.
As noise increases, the level of firing rates also rises, and the Softplus activation function approximates the response firing activity driven by current with Gaussian white noise added.
Softplus units only represent a single level of noise that, for example, the dotted line in Figure~\ref{Fig:physics} is drawn when $s_I=1$.

\begin{table}[bt]
	\centering
	\caption{\label{tbl:pynnConfig}Parameter setting for the current-based LIF neurons using PyNN.}
	\bgroup
	\def\arraystretch{1.4}
	\begin{tabular}{c c c}
		Parameters & Values & Units \\
		\hline
		cm & 0.25 & nF	\\
		tau\_m & 20.0 & ms\\
		tau\_refrac & 1.0 & ms\\
		v\_reset & -65.0 & mV\\
		v\_rest & -65.0 & mV\\
		v\_thresh & -50.0 & mV\\
		i\_offset & 0.1 & nA\\
		\hline
	\end{tabular}
	\egroup
\end{table}

%

Although the use of Siegert function opened the door for modelling response activities of LIF neurons as activation function used in ANNs, there are several drawbacks of this method:
\begin{itemize}
	\item most importantly, the numerical analysis on an LIF response function is far from accurate in practice. ``Practice'' in the paper means SNN simulations using LIF neurons.
	Thus the inaccurate model generates error between the estimation and the practical response firing rate.
	This issue will be addressed in Section~\ref{subsec:practice}.

	\item the high complexity of the Siegert function causes much longer training time and more energy, let alone the high-cost computation on its derivative.
	\item the training uses Siegert function to estimate the output firing rate in the forward path, but uses the derivative of sigmoid function during back propagation for lower complexity.
	However, due to the difference between Siegert and sigmoid functions, error is introduced. 
	\item neurons have to fire at high frequency (higher than half of the maximum firing rate) to represent the activation of a sigmoid unit; as a result the network activity results in high power dissipation.
	\item better learning performance has been reported using ReLU, so modelling ReLU-like activation function for spiking neurons is needed.  
\end{itemize}

Therefore, we propose the Noisy Softplus function which provides solutions to the drawbacks of Siegert unit.

\subsection{Mismatch of Siegert Function to Practice }
\label{subsec:practice}

\begin{figure}[tbp!]
	\centering
	\begin{subfigure}[t]{0.49\textwidth}
		\includegraphics[width=\textwidth]{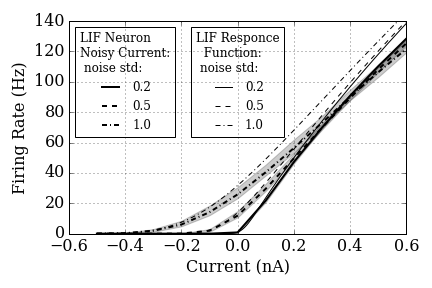}
		\caption{Current sampled at $dt$=1~ms.}
	\end{subfigure}
	\begin{subfigure}[t]{0.49\textwidth}
		\includegraphics[width=\textwidth]{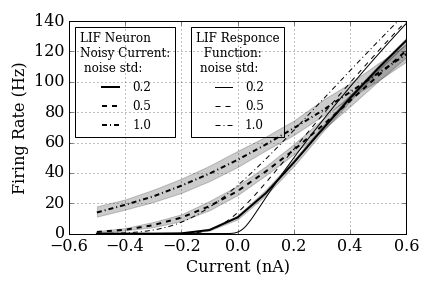}
		\caption{Current sampled at $dt$=10~ms.}
	\end{subfigure}
	\caption{Recorded response firing rate of a LIF neuron driven by \textit{NoisyCurrentSource} sampled at every (a) 1~ms and (b) 10~ms. Averaged firing rates of 10 simulation trails are shown in bold lines, and the grey colour fills the range between the minimum to maximum of the firing rates. The analytical LIF response function, Siegert formula (Equation~(\ref{equ:siegert})), is drawn in thin lines (shown in Figure~\ref{Fig:physics}) to compare with the practical simulation.}
	\label{Fig:current}
\end{figure}

\begin{figure}[tbp!]
	\centering
	\makebox[0.43\textwidth]{$dt$=1~ms} \makebox[0.43\textwidth]{$dt$=10~ms} \par
	\begin{subfigure}[t]{0.43\textwidth}
		\includegraphics[width=\textwidth]{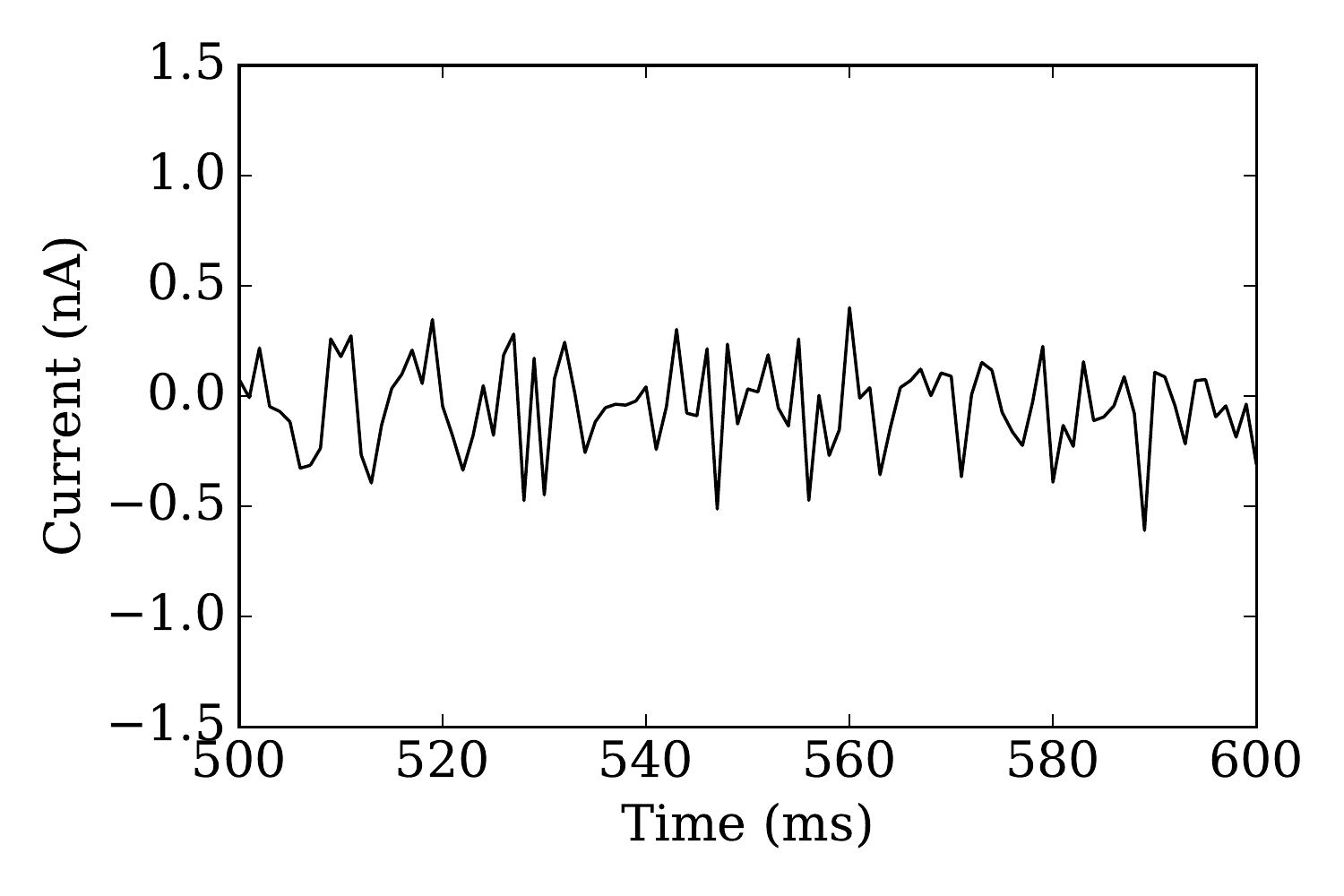}
		\caption{Current sampled at $dt$=1~ms.}
	\end{subfigure}
	\begin{subfigure}[t]{0.43\textwidth}
		\includegraphics[width=\textwidth]{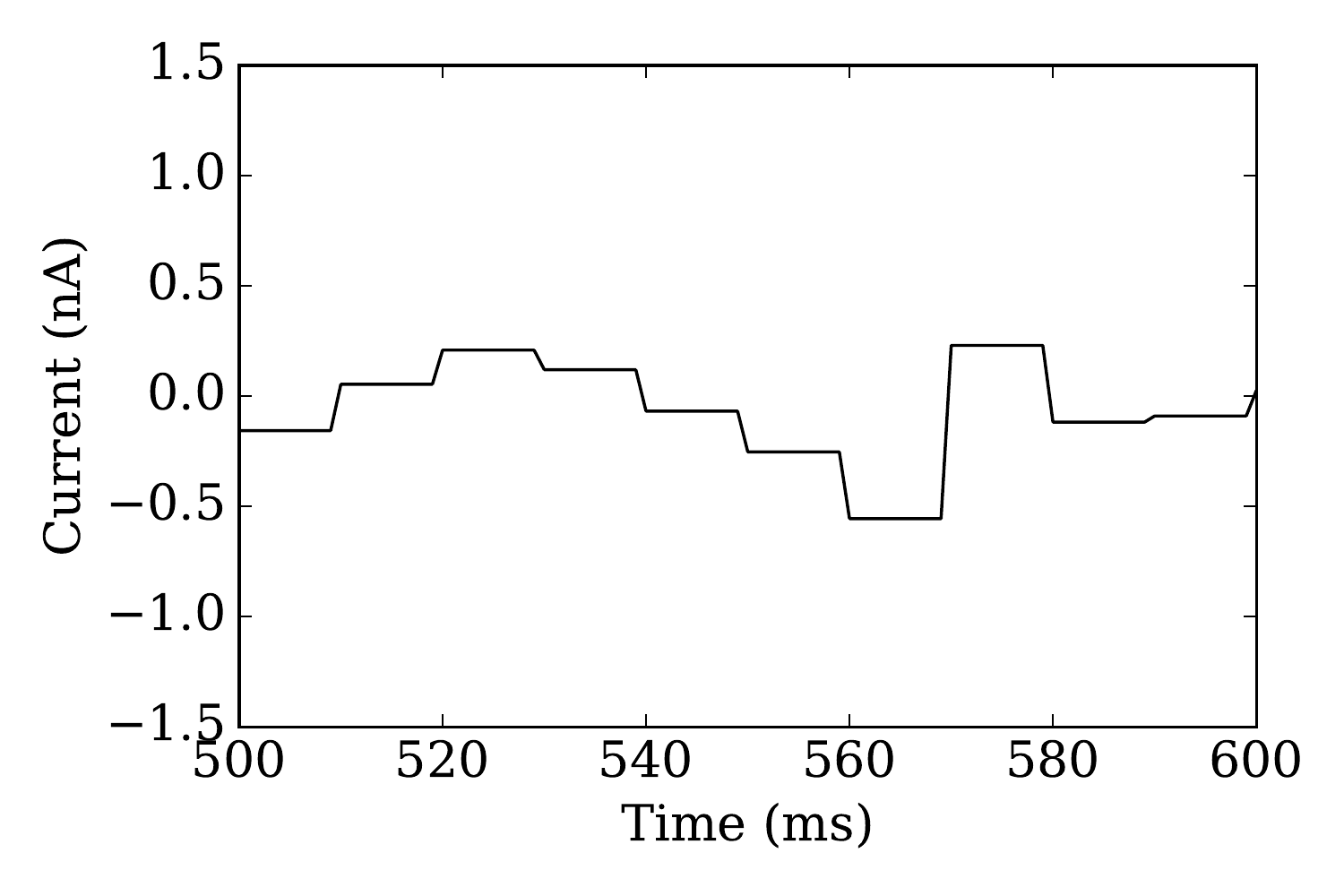}
		\caption{Current sampled at $dt$=10~ms.}
	\end{subfigure}\\
	\begin{subfigure}[t]{0.43\textwidth}
		\includegraphics[width=\textwidth]{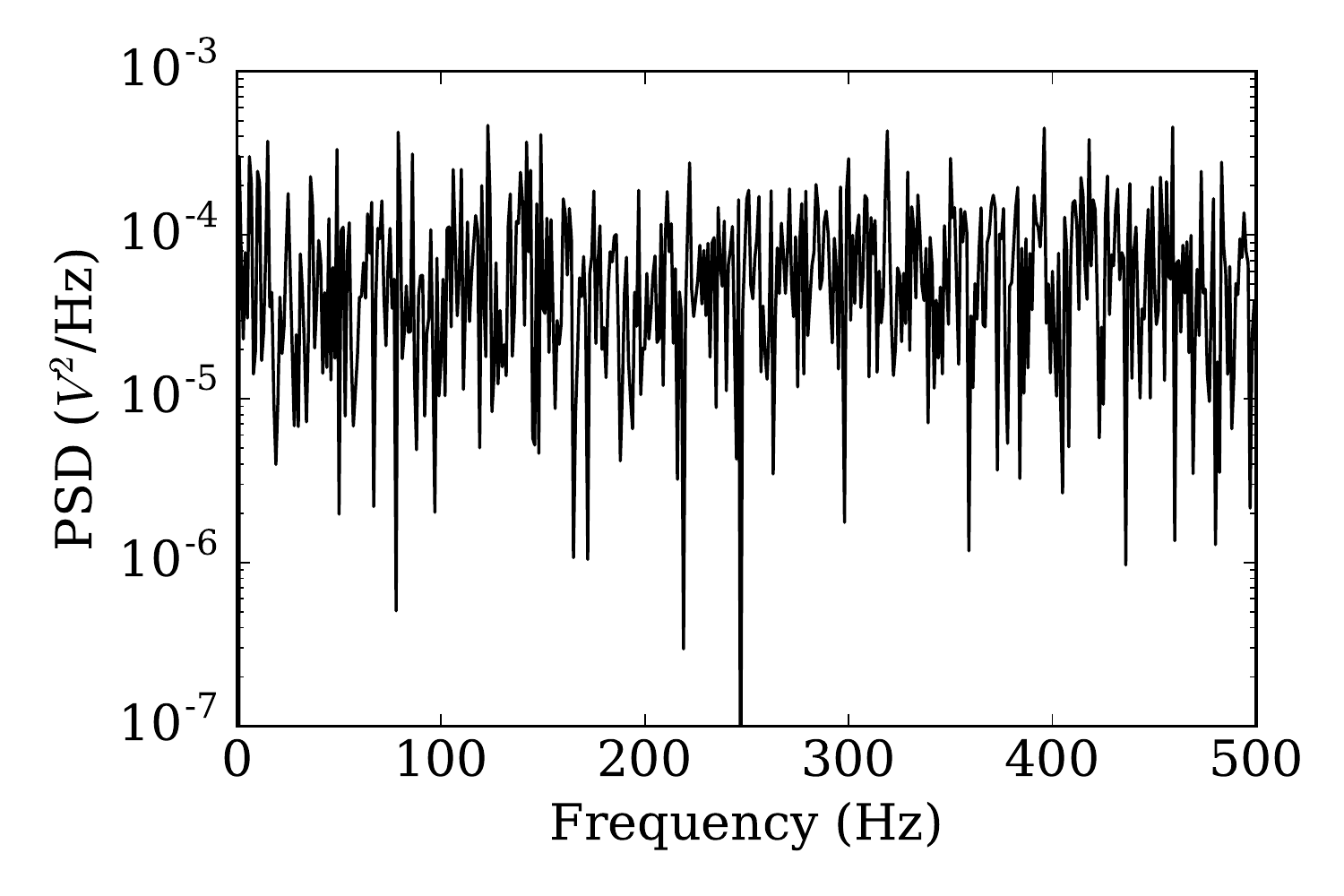}
		\caption{Spectrum analysis of (a).}
	\end{subfigure}
	\begin{subfigure}[t]{0.43\textwidth}
		\includegraphics[width=\textwidth]{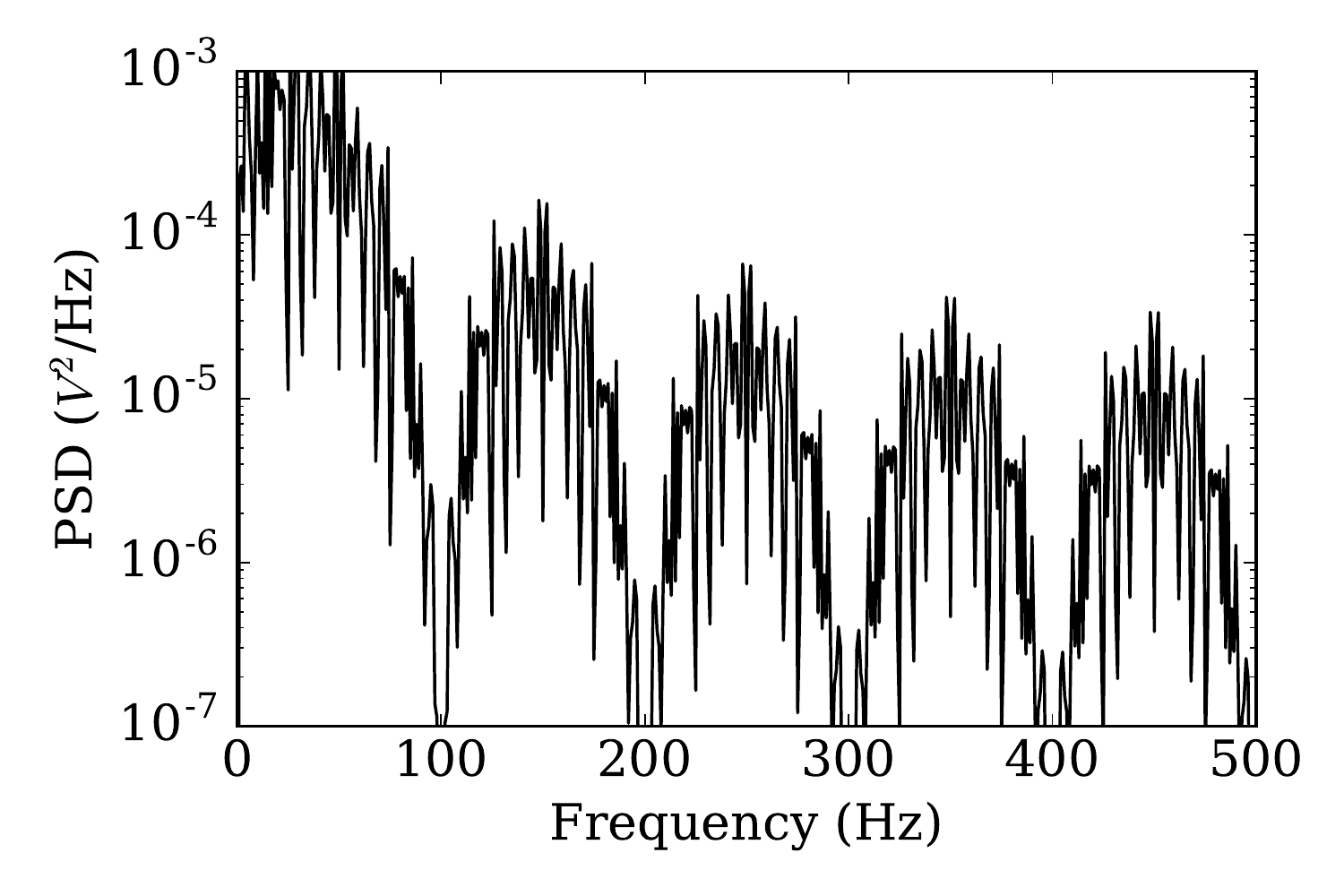}
		\caption{Spectrum analysis of (b).}
	\end{subfigure}\\
	\begin{subfigure}[t]{0.43\textwidth}
		\includegraphics[width=\textwidth]{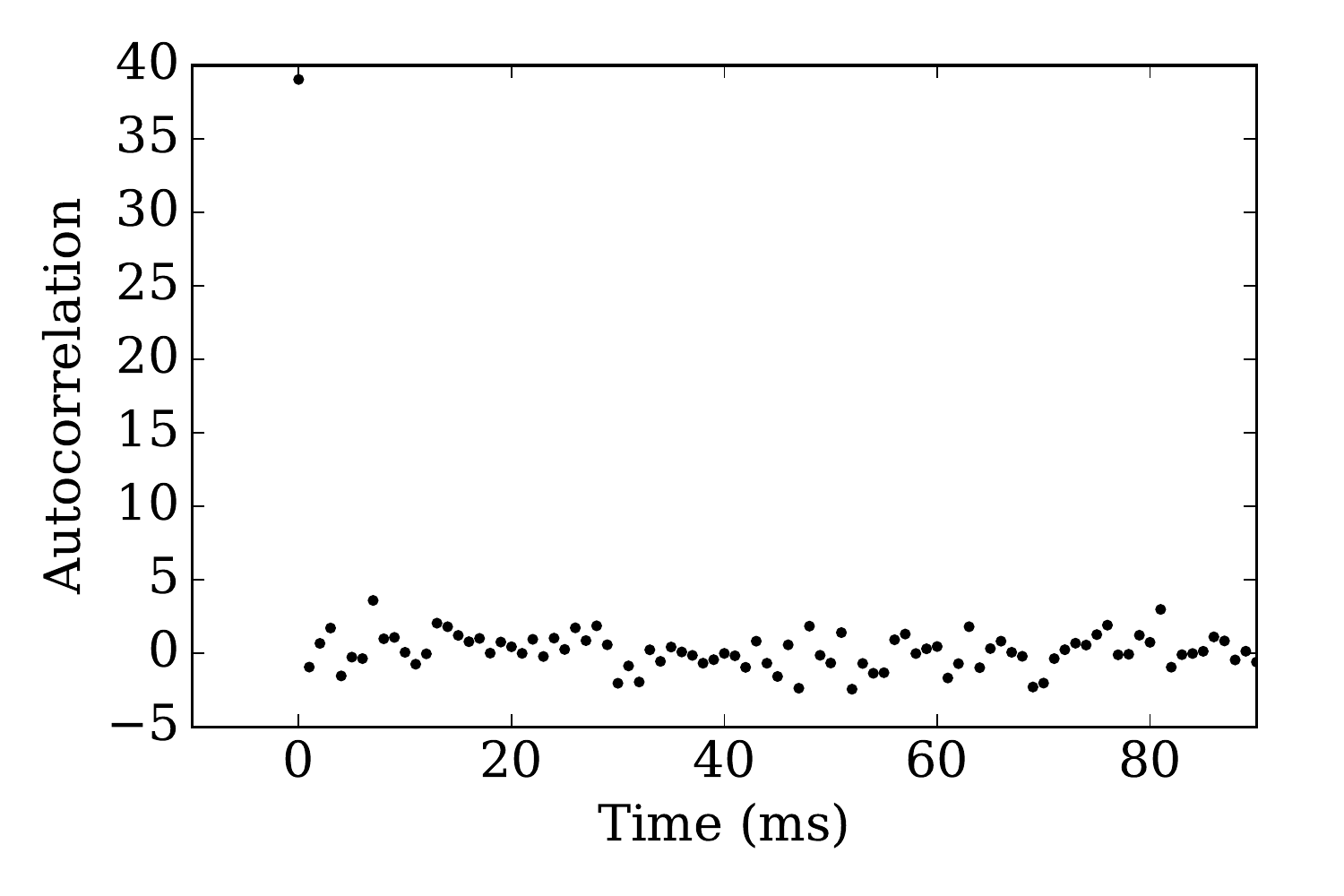}
		\caption{Autocorrelation of (a).}
	\end{subfigure}
	\begin{subfigure}[t]{0.43\textwidth}
		\includegraphics[width=\textwidth]{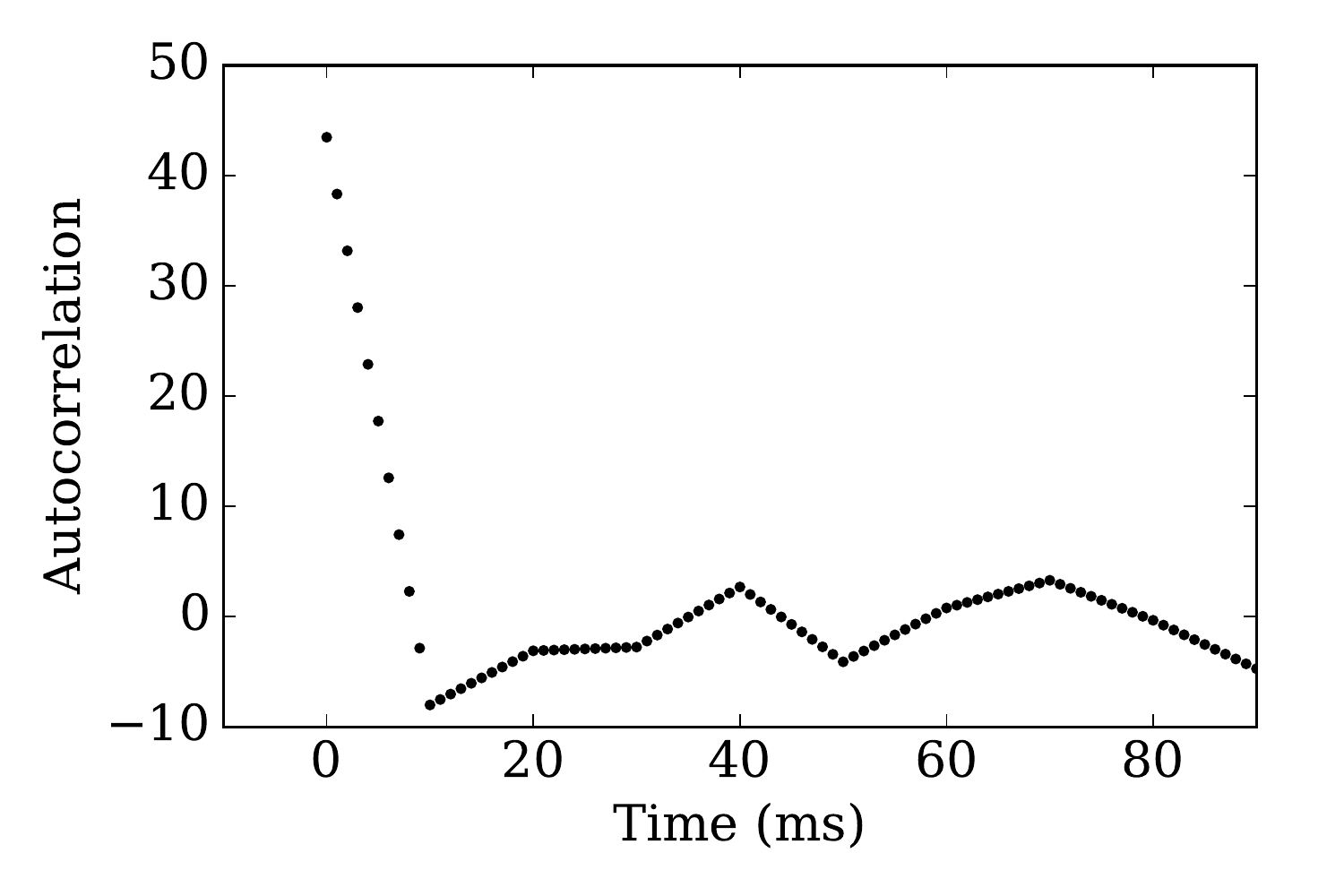}
		\caption{Autocorrelation of (b).}
	\end{subfigure}\\
	\begin{subfigure}[t]{0.43\textwidth}
		\includegraphics[width=\textwidth]{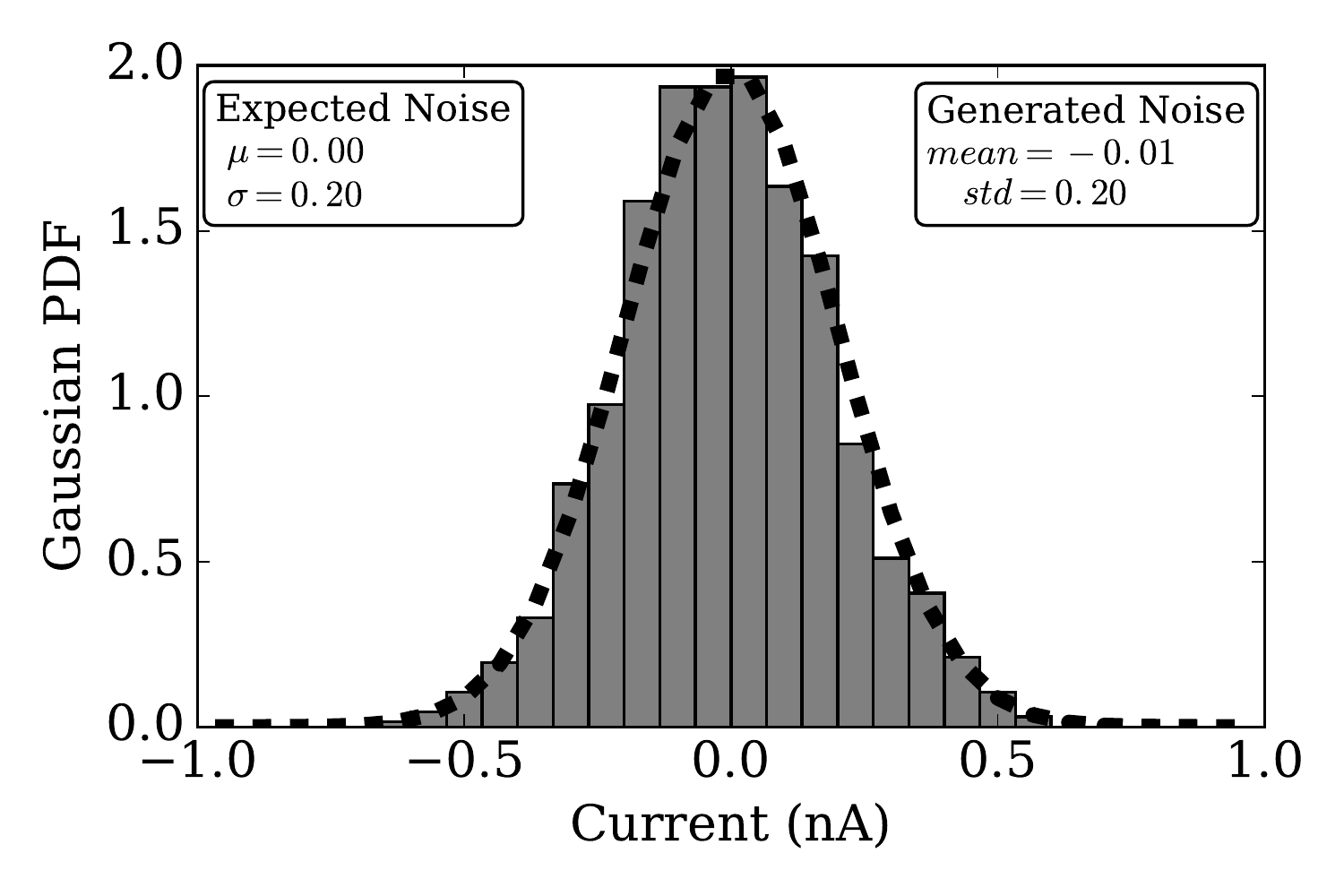}
		\caption{Distribution of samples of (a).}
	\end{subfigure}
	\begin{subfigure}[t]{0.43\textwidth}
		\includegraphics[width=\textwidth]{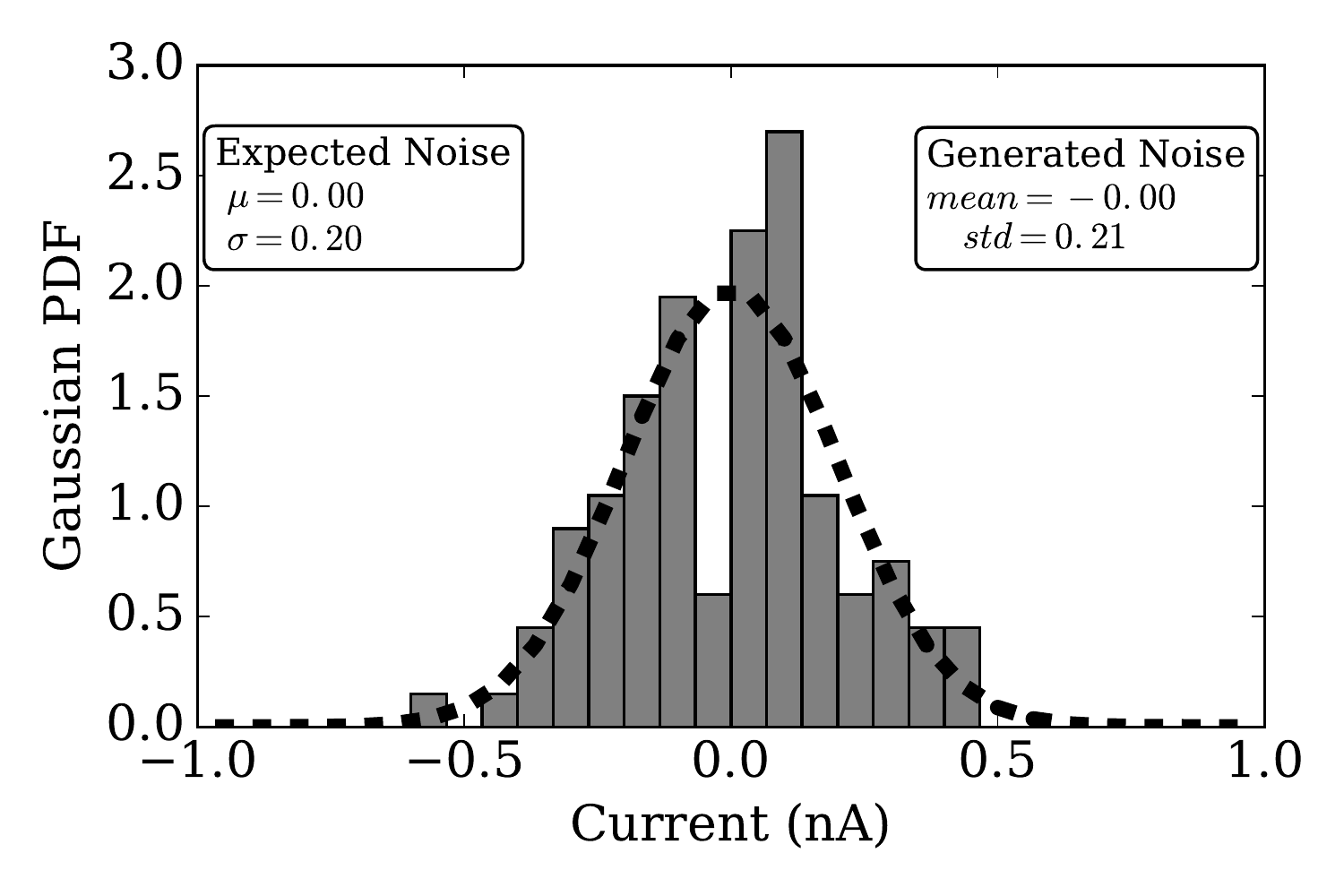}
		\caption{Distribution of samples of (b).}
	\end{subfigure}
	\caption{\textit{NoisyCurrentSource} samples noisy currents from Gauss distribution at every 1~ms (left) and 10~ms (right). The signals are shown in time domain in (a) and (b), and spectrum domain in (c) and (d). The autocorrelation of both current signals are shown in (e) and (f). The distribution of the discrete samples are plotted in bar chart to compare with PDF of the original Gauss distribution, shown in (g) and (h).}
	\label{Fig:lif_curr}
\end{figure}

To verify the response function in practice, simulation tests were carried out using PyNN~\cite{davison2008pynn} to compare with the analytical results.
The noisy current was produced by \textit{NoisyCurrentSource} in PyNN which is a constant current of $m_I$ added to a Gaussian white noise of zero mean and $s_I^2$ variance.
The noise was drawn from the Gaussian distribution in a time resolution of $dt$.
We chose $dt=1$~ms which is the finest resolution for common SNN simulators, and $dt=10$~ms for comparison.
For a given pair of $m_I$ and $s_I^2$, a noisy current was injected into a current-based LIF neuron for 10~s, and the output firing rate was the average over 10 trials.

The curves in Figures~\ref{Fig:current} illustrate output firing rate driven by different level of noise
The differences relative to the analytical results (thin lines) is due to the time resolution, $dt$, of the \textit{NoisyCurrentSource}.
The sampled signals are shown in Figures~\ref{Fig:lif_curr}~(a) and (b).
The discrete sampling of the noisy current introduces time step correlation to the white noise, shown in Figures~\ref{Fig:lif_curr}~(e) and (f), where the value remains the same within a time step $dt$.
Although both current signals followed the same Gauss distribution, see Figures~\ref{Fig:lif_curr}~(g) and (h), the current is a white noise when $dt=1$~ms, but a coloured noise, e.g. increased Power Spectral Density~(PSD) at lower frequency, when $dt=10$~ms, see Figures~\ref{Fig:lif_curr}~(c) and (d).
Thus the Siegert formula, Equation~(\ref{equ:siegert}), can only approximate the LIF response of noisy current with white noise.

\begin{figure}[tbp!]
	\centering
	\begin{subfigure}[t]{0.49\textwidth}
		\includegraphics[width=\textwidth]{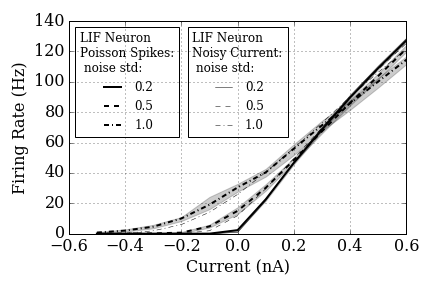}
		\caption{$\tau_{syn}$=1~ms.}
	\end{subfigure}
	\begin{subfigure}[t]{0.49\textwidth}
		\includegraphics[width=\textwidth]{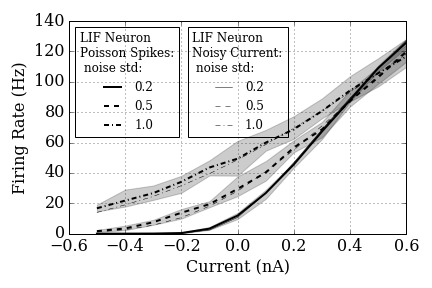}
		\caption{$\tau_{syn}$=10~ms.}
	\end{subfigure}
	\caption{Recorded response firing rate of a LIF neuron driven by synaptic current with two synaptic time constants: (a) $\tau_{syn}$=10~ms and (b) $\tau_{syn}$=10~ms. Averaged firing rates of 10 simulation trails are shown in bold lines, and the grey colour fills the range between the minimum to maximum of the firing rates. The firing rates recorded driven by \textit{NoisyCurrentSource}, are drawn in thin lines (shown in Figure~\ref{Fig:lif_curr}) to compare with.}
	\label{Fig:spike_curr}
\end{figure}

\begin{figure}[tbp!]
	\centering
	\makebox[0.43\textwidth]{$\tau_{syn}$=1~ms} \makebox[0.43\textwidth]{$\tau_{syn}$=10~ms} \par
	\begin{subfigure}[t]{0.43\textwidth}
		\includegraphics[width=\textwidth]{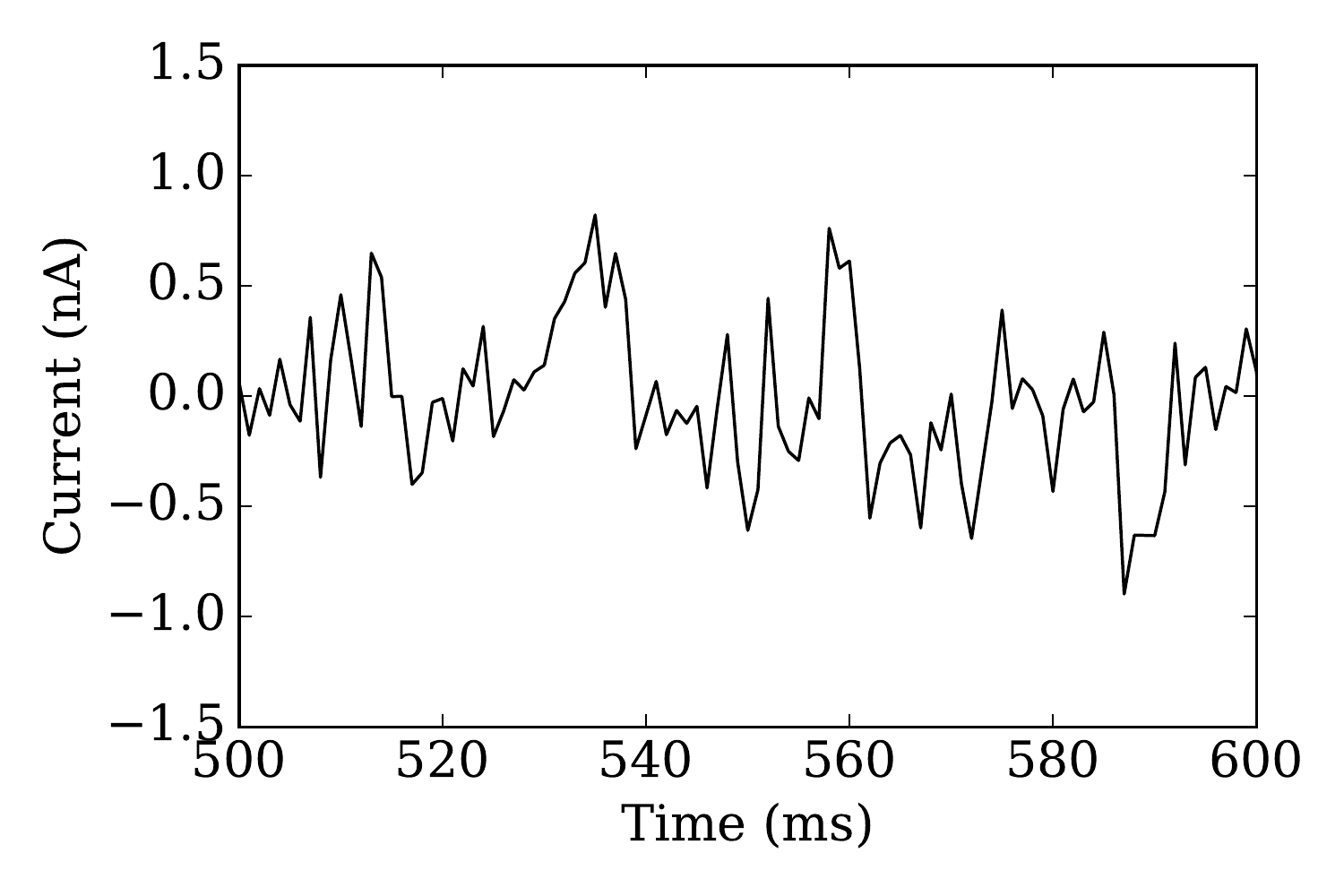}
		\caption{Current sampled at $dt$=1~ms.}
	\end{subfigure}
	\begin{subfigure}[t]{0.43\textwidth}
		\includegraphics[width=\textwidth]{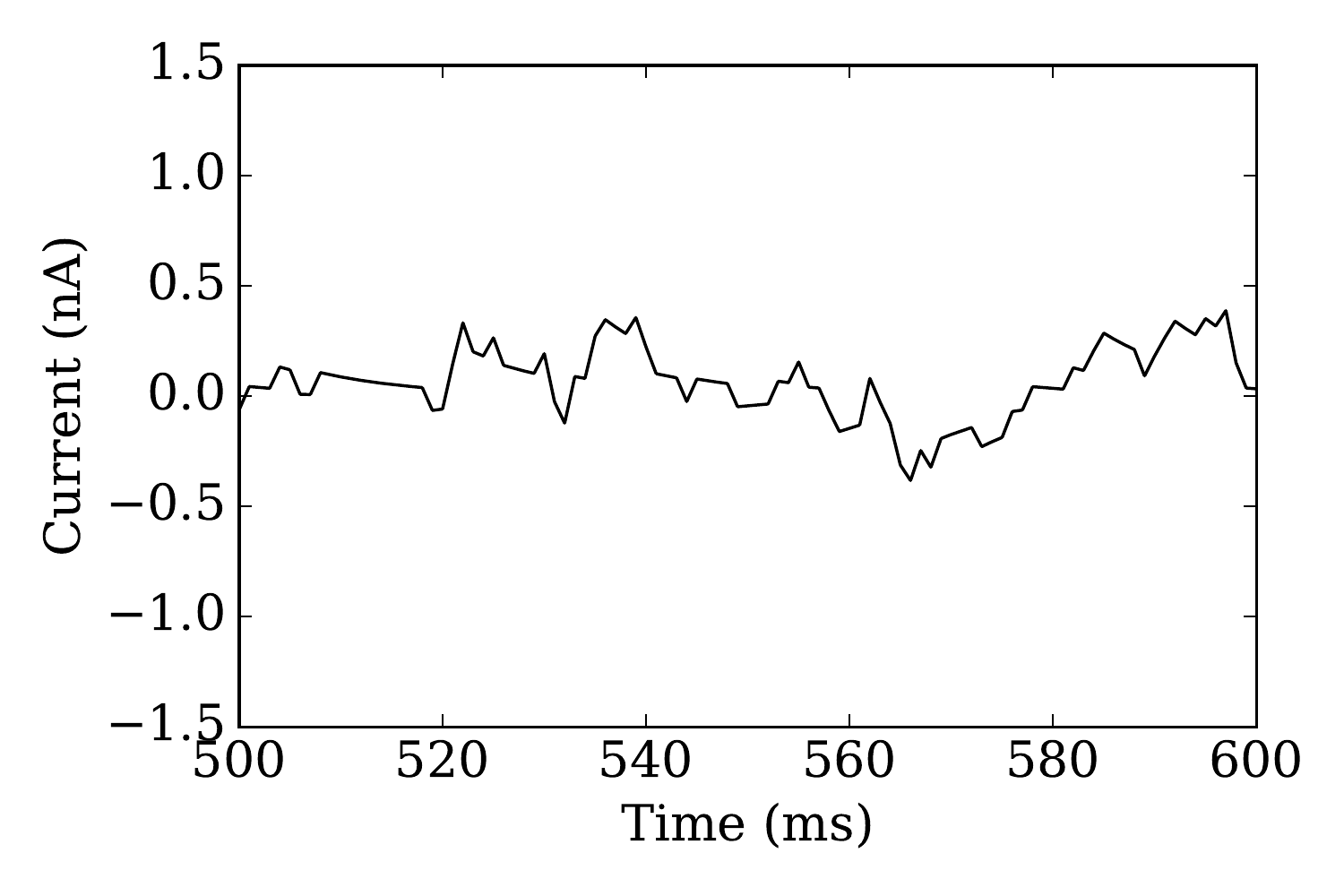}
		\caption{Current sampled at $dt$=10~ms.}
	\end{subfigure}\\
	\begin{subfigure}[t]{0.43\textwidth}
		\includegraphics[width=\textwidth]{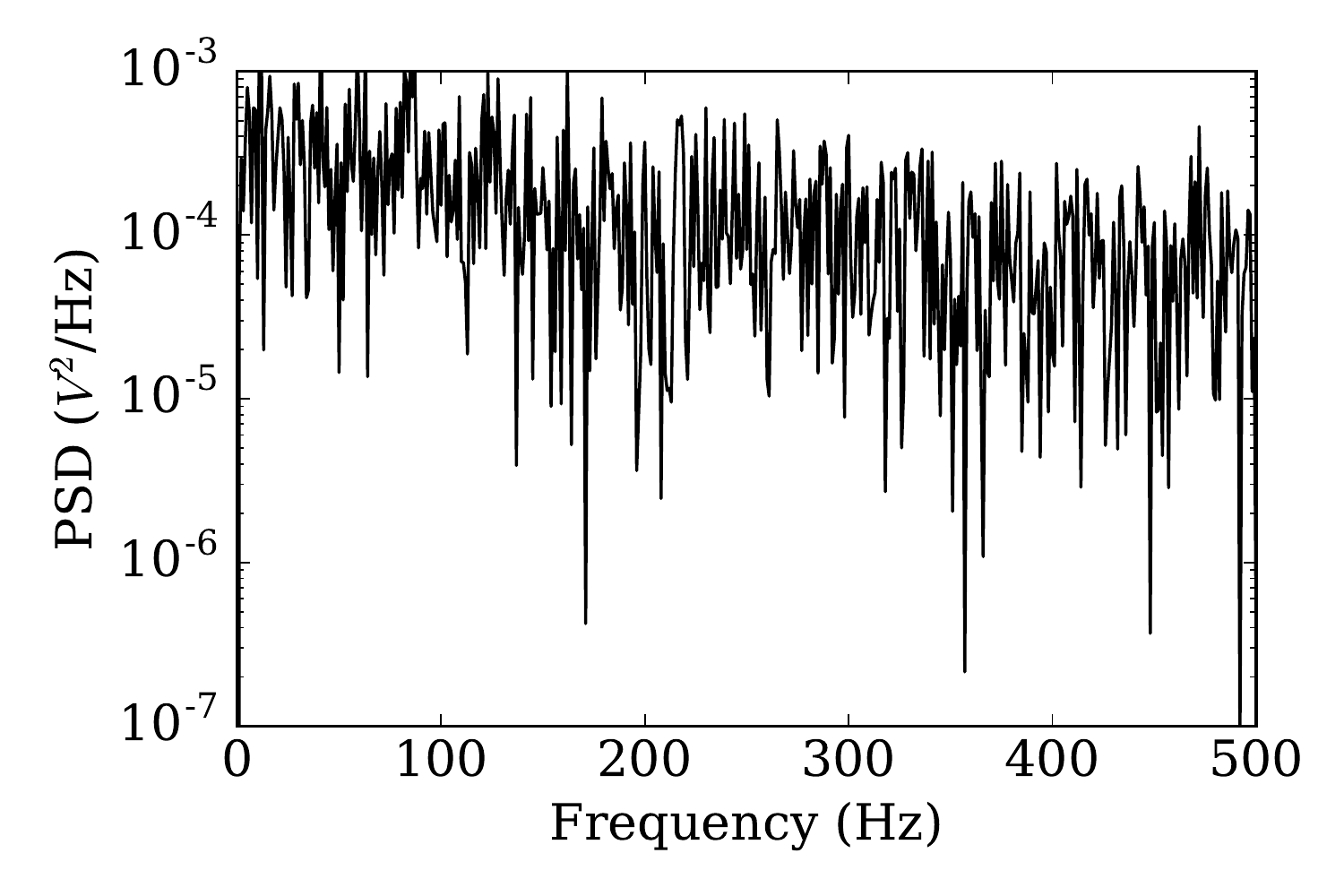}
		\caption{Spectrum analysis of (a).}
	\end{subfigure}
	\begin{subfigure}[t]{0.43\textwidth}
		\includegraphics[width=\textwidth]{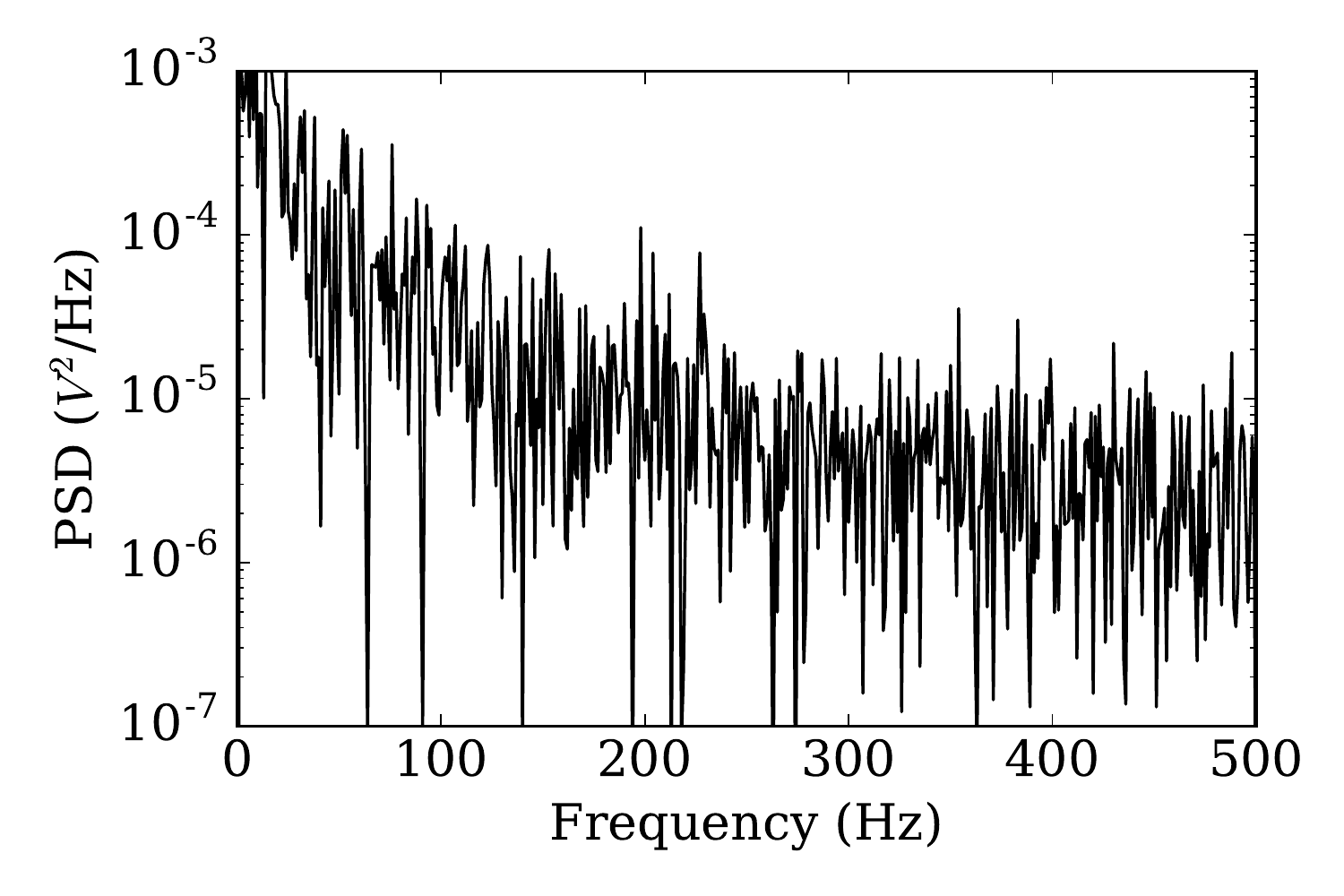}
		\caption{Spectrum analysis of (b).}
	\end{subfigure}\\
	\begin{subfigure}[t]{0.43\textwidth}
		\includegraphics[width=\textwidth]{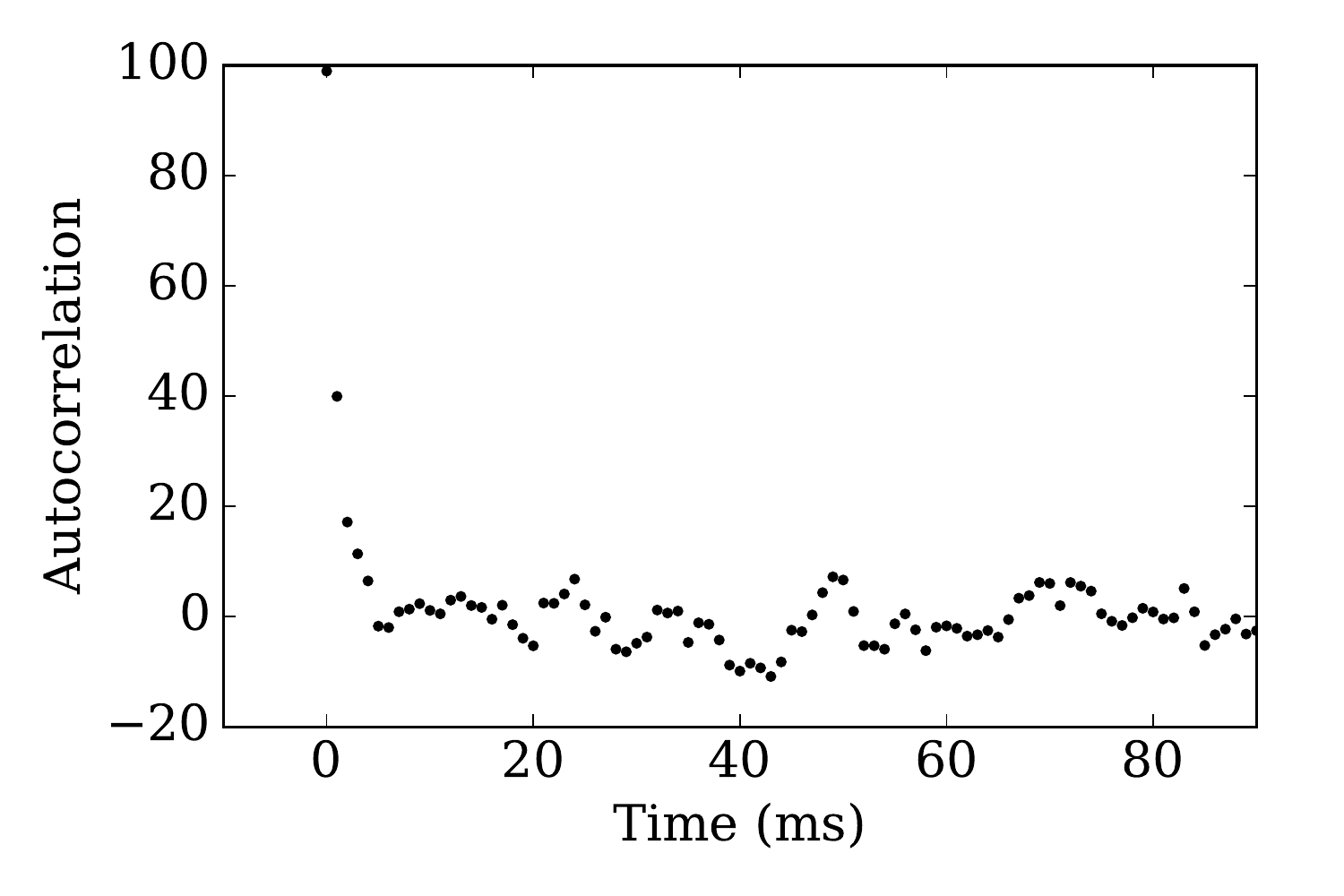}
		\caption{Autocorrelation of (a).}
	\end{subfigure}
	\begin{subfigure}[t]{0.43\textwidth}
		\includegraphics[width=\textwidth]{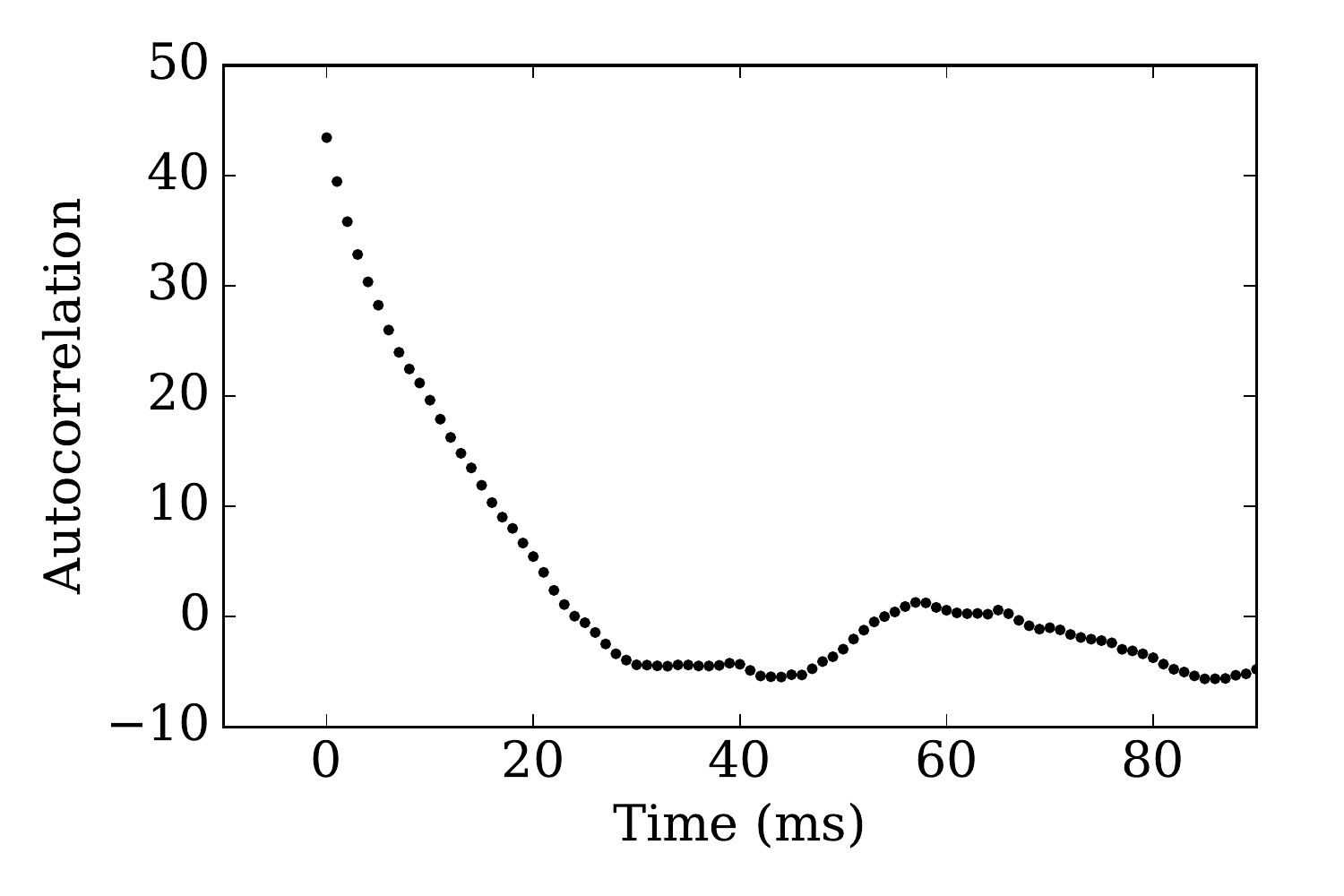}
		\caption{Autocorrelation of (b).}
	\end{subfigure}\\
	\begin{subfigure}[t]{0.43\textwidth}
		\includegraphics[width=\textwidth]{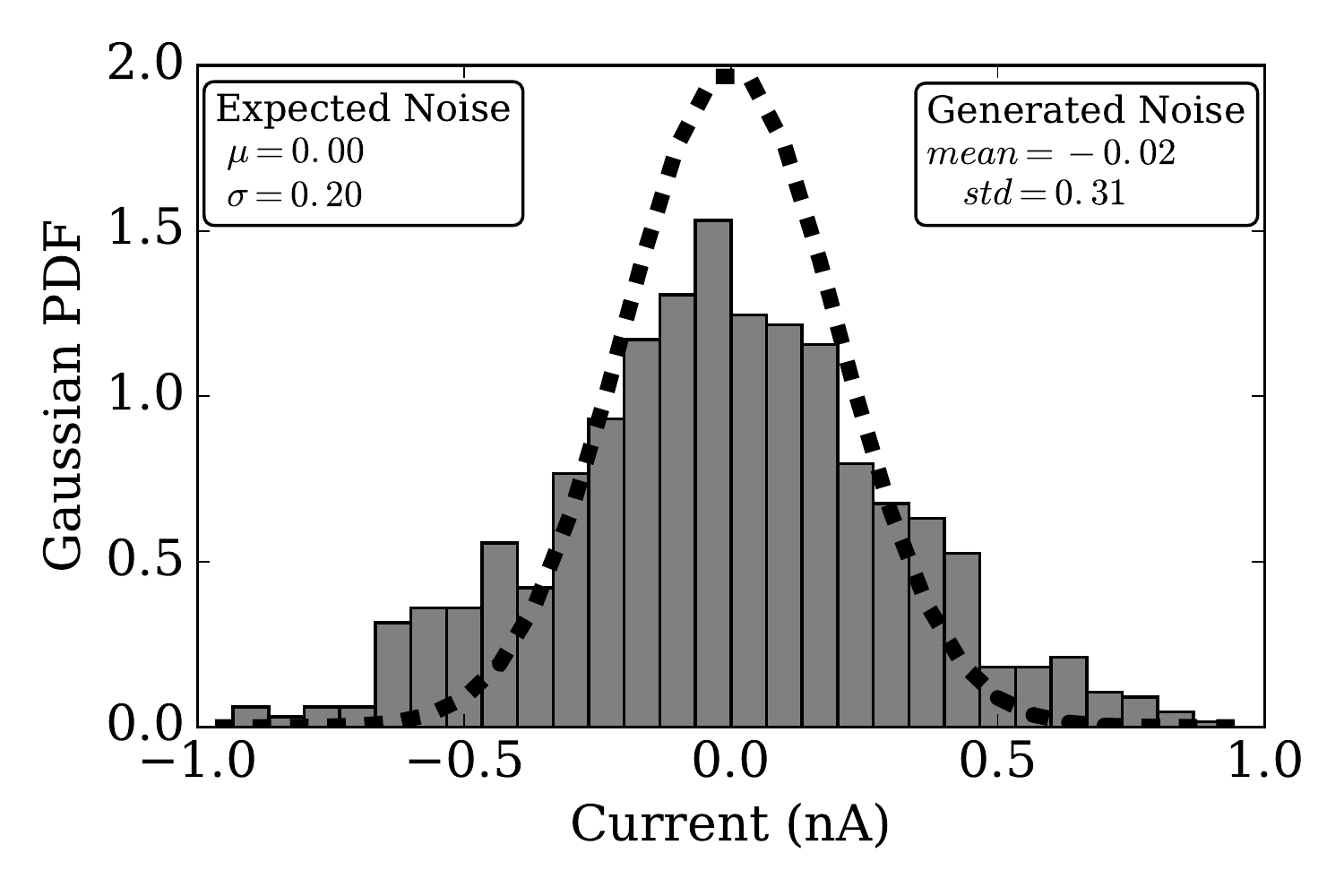}
		\caption{Distribution of samples of (a).}
	\end{subfigure}
	\begin{subfigure}[t]{0.43\textwidth}
		\includegraphics[width=\textwidth]{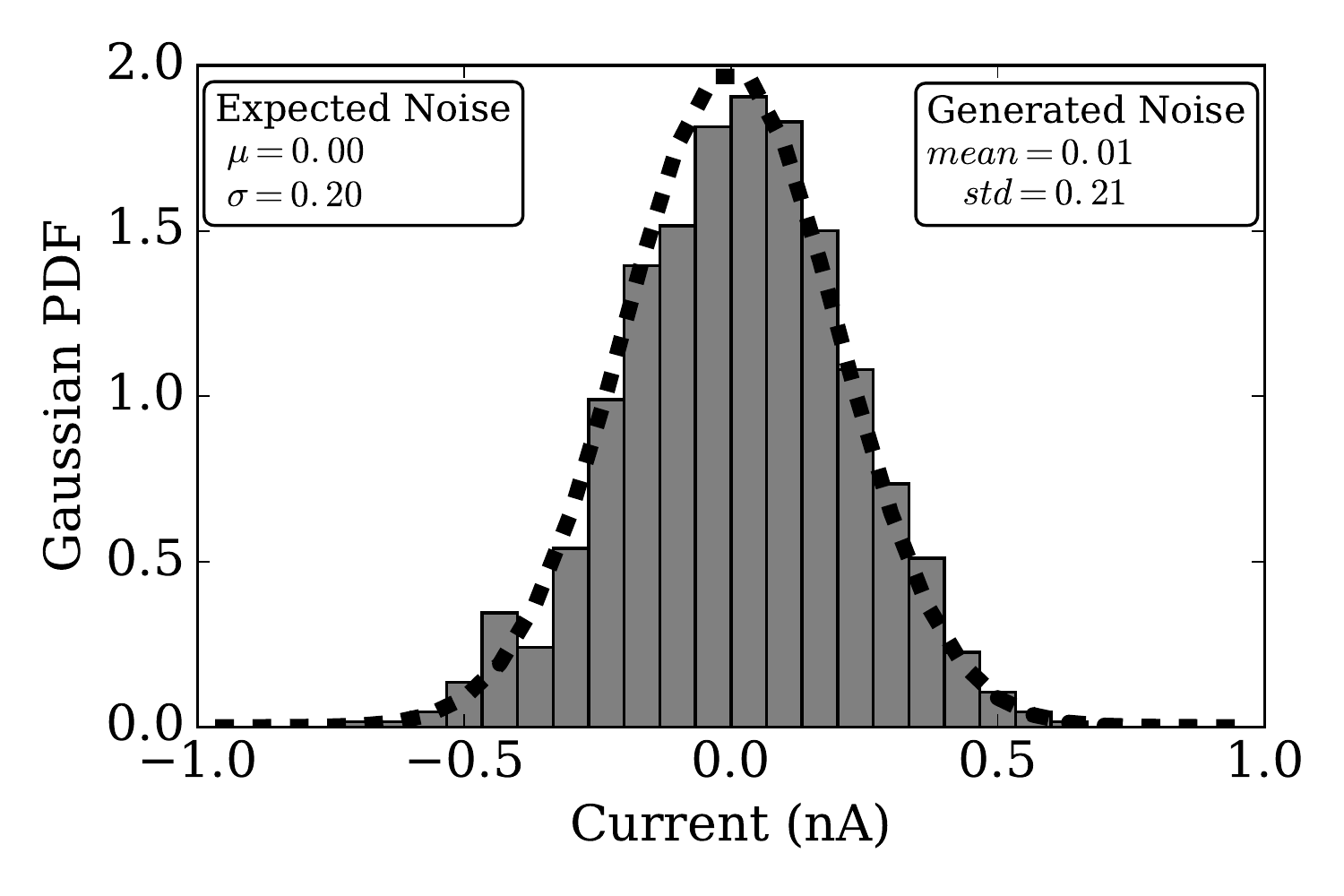}
		\caption{Distribution of samples of (b).}
	\end{subfigure}
	\caption{Noisy currents generated by 100 Poinsson spike trains to a LIF neuron with synaptic time constant $\tau_{syn}$=1~ms (left) and $\tau_{syn}$=10~ms (right). The currents are shown in time domain in (a) and (b), and spectrum domain in (c) and (d). The autocorrelation of both current signals are shown in (e) and (f). The distribution of the generated samples are plotted in bar chart to compare to the expected Gauss distribution, shown in (g) and (h).}
	\label{Fig:lif_pois}
\end{figure}

A more realistic simulation of a noisy current can be generated by 100 Poisson spike trains, 
where the mean and variance of the current are given by:
\begin{equation}
m_I = \tau_{syn}\sum_i w_i\lambda_{i}~, ~s_I^2=\frac{1}{2}\tau_{syn}\sum_i w_i^2\lambda_{i}~,
\label{equ:distr}
\end{equation}
where $\tau_{syn}$ is the synaptic time constant, and each Poisson spike train connects to the neuron with a strength of $w_i$ and a firing rate of $\lambda_i$.
Two populations of Poisson spike sources, for excitatory and inhibitory synapses respectively, were connected to a single LIF neuron to mimic the noisy currents.
The firing rates of the Poisson spike generators were determined by the given $m_I$ and $s_I$.
Figures~\ref{Fig:spike_curr} illustrate the recorded firing rates responding to the Poissoin spike trains compared to the mean firing rate driven by \textit{NoisyCurrentSource} in Figure~\ref{Fig:lif_curr}.
Note that the estimation of LIF response activity requires noisy current with white noise, however
in practice the release of neurotransmitter takes time ($\tau_{syn} >> 0$) and the synaptic current the decays exponentially $I_{syn} = I_0 e^{\frac{-t}{\tau_{syn}}}$.
Figures~\ref{Fig:lif_pois}~(a) and (b) show two examples of synaptic current of 0~nA mean and 0.2 standard deviation driven by 100 neurons firing at the same rate and with the same synaptic strength (half excitatory, half inhibitory), but of different synaptic time constant.
Therefore, the current at any time t during decaying period is dependant to the value at previous time step, which makes the synaptic current a coloured noise, see Figures~\ref{Fig:lif_pois}~(c) and (d).

We observe in Figure~\ref{Fig:spike_curr}~(a) that the response firing rate to synaptic current is higher than the \textit{NoisyCurrentSource} for all the 10 trials.
This is caused by the coarse resolution (1~ms) of the spikes, thus the standard deviation of the current is larger than 0.2, shown in Figure~\ref{Fig:lif_pois}~(g);
and the $\tau_{syn}$, even short as 1~ms, adds coloured noise instead of white noise to the current.
However, Figure~\ref{Fig:lif_pois}~(h) shows a similar firing rate of both the synaptic driven current and the \textit{NoisyCurrentSource}, since both of the current signals have similar distribution and time correlation.
Nevertheless, the analytical response function, Siegert formula, cannot approximate either of the practical simulations.


\subsection{Noisy Softplus}
\begin{figure}[thb!]
	\centering
	\begin{subfigure}[t]{0.6\textwidth}
		\includegraphics[width=\textwidth]{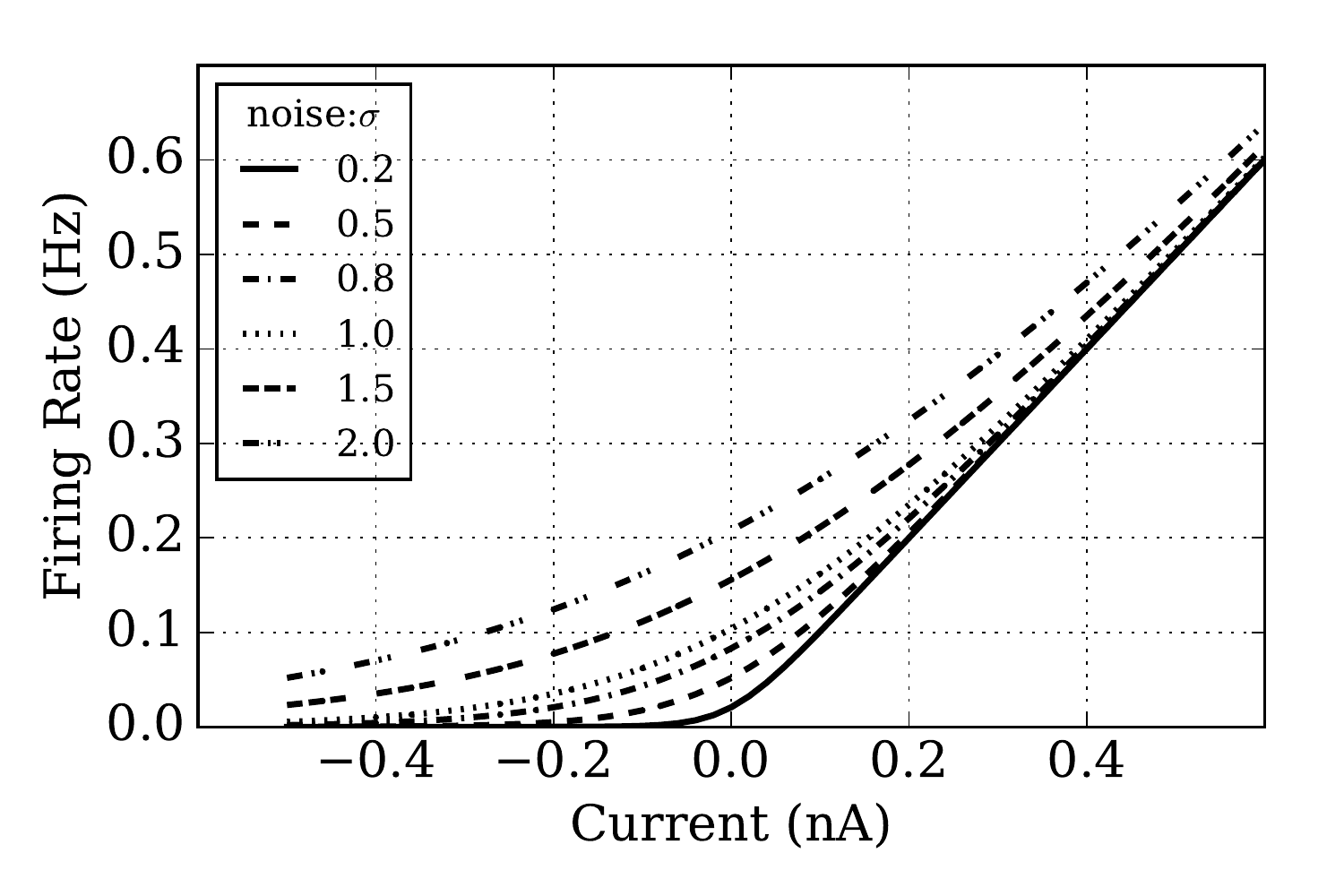}
		\caption{Noisy Softplus}
	\end{subfigure}\\
	\begin{subfigure}[t]{0.7\textwidth}
		\includegraphics[width=\textwidth]{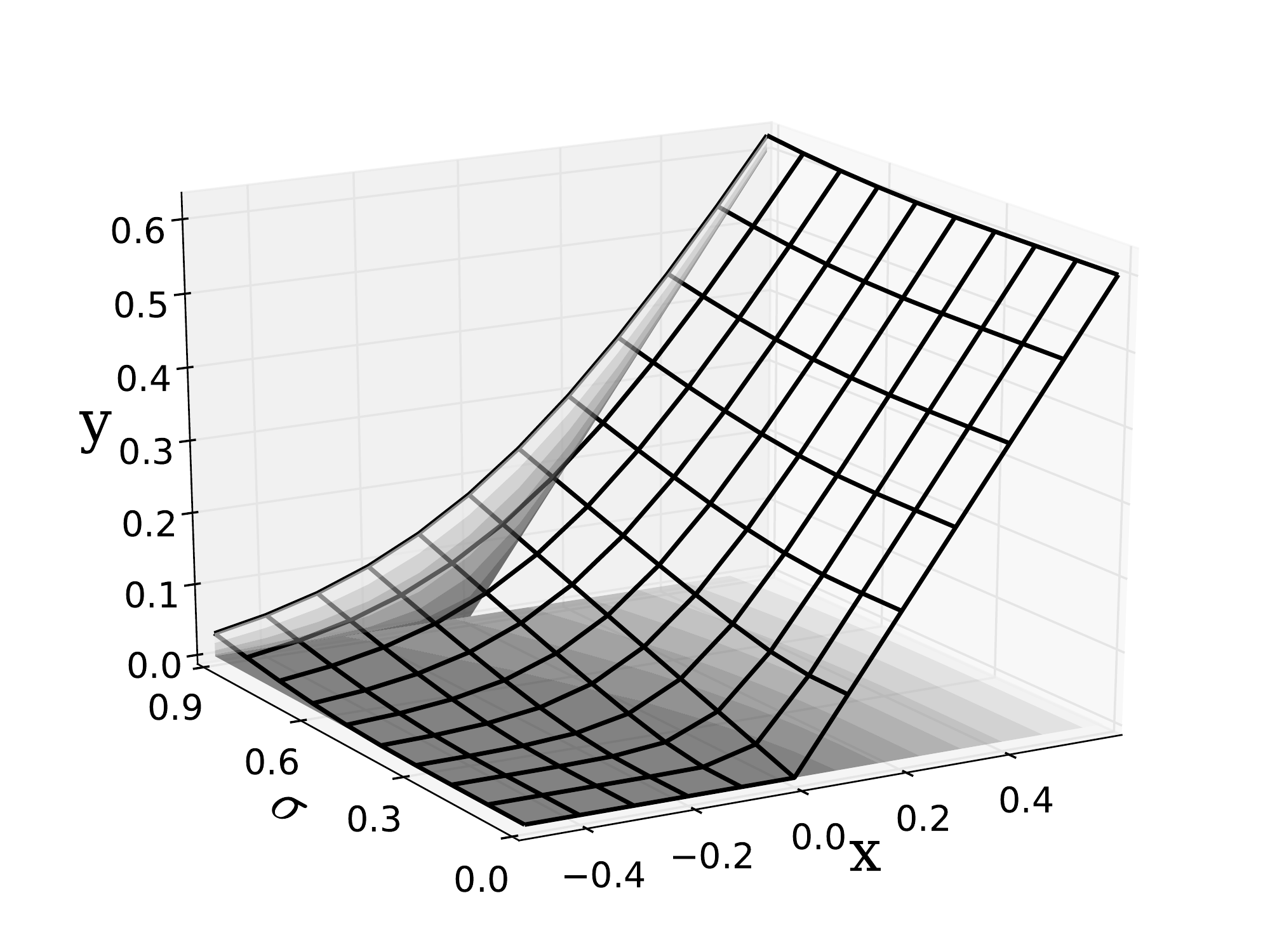}
		\caption{Noisy Softplus in 3D}
	\end{subfigure}
	\caption{
		Noisy Softplus fits to the response function of the LIF neuron.
		Noisy Softplus in (a) curve sets and (b) 3D.}
	\label{fig:nsp}
\end{figure}

Due to the limited time resolution of common SNN simulators and the time taken for neurotransmitter release, $\tau_{syn}$, mismatches exist between the analytical response function, the Siegert formula, and practical neural activities.
Consequently, a unified activation function is required to model the practical responses of a LIF neuron.
Inspired by the set of response functions triggered by different levels of noise, we propose the Noisy Softplus activation function:
\begin{equation}
y = f_{ns}(x, \sigma) = k \sigma \log [1 + \exp(\frac{x}{k \sigma})],
\label{equ:nsp}
\end{equation}
where $x$ refers to the mean current, $y$ indicates the strength of the output firing rate, $\sigma$ plays an important role to define the noise level, and $k$, which is determined by the neuron parameters, controls the shape of the curves.
Note that the novel activation function we propose contains two parameters, the current and its noise; both are naturally obtained in spiking neurons.
Figure~\ref{fig:nsp} shows the activation function in curve sets corresponding to different discrete noise levels, and in a 3D plot.

The derivative is the logistic function scaled by $k\sigma$:
\begin{equation}
\frac{\partial f_{ns}(x,\sigma)}{\partial x} = \frac{1}{1+exp(-\frac{x}{k\sigma})}~~,
\label{equ:logist}
\end{equation}	
which could be easily applied to back propagation in any network training.
However, such a derivative function of low complexity does not present in Siegert function.

The activation function can be scaled up by a factor, $S$, to represent the firing rate $\lambda$ of a LIF neuron driven by a noisy current $x$.
\begin{equation}
\begin{aligned}
\lambda_{out} &\simeq f_{ns}(x, \sigma) \times S\\
&=k \sigma \log [1 + \exp(\frac{x}{k \sigma})] \times S
\end{aligned}
\label{equ:fit}
\end{equation}	

\begin{figure}
	\centering
	\begin{subfigure}[t]{0.49\textwidth}
		\includegraphics[width=\textwidth]{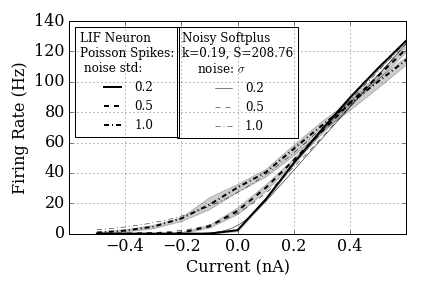}
		\caption{$\tau_{syn}$=1~ms}
	\end{subfigure}
	\begin{subfigure}[t]{0.49\textwidth}
		\includegraphics[width=\textwidth]{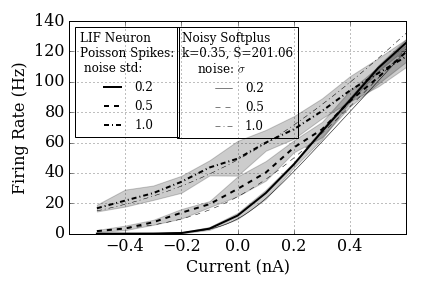}
		\caption{$\tau_{syn}$=10~ms}
	\end{subfigure}
	\caption{Noisy Softplus fits to the response firing rates of LIF neurons.}
	\label{Fig:nsptau1}
\end{figure}

Noisy Softplus fits well to the practical response firing rate of the LIF neuron with suitable calibration of $k$ and $S$, see Figure~\ref{Fig:nsptau1}.
The parameter pair of $(k, S)$ is curve-fitted with the triple data points of $(\lambda, x, \sigma)$.
The fitted parameter was set to $(k, S)=(0.19,208.76)$ for the practical response firing rate driven by synaptic noisy current with $\tau_{syn}=1$~ms and was set to $(k, S)=(0.35,201.06)$ when $\tau_{syn}=10$~ms.
The calibration currently is conducted by linear least squares regression; numerical analysis is considered however for future work to express the factors with biological parameters of a LIF neuron.

\section{ANN-Trained SNNs}
\label{sec:ann_train_snn}
We have discussed modelling response firing activity of a LIF neuron with a unified activation function, Noisy Softplus.
However the demonstration of mapping the physical activity to numerical ANN calculations is still required to train the layered-up deep network.

\subsection{Equivalent Input and Output}

\begin{figure}[bt]
	\centering
	\includegraphics[width=0.7\textwidth]{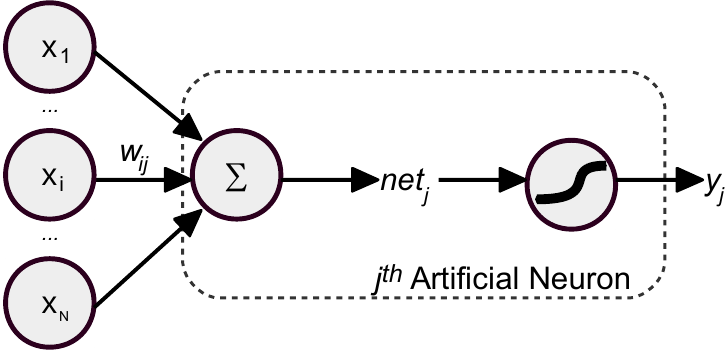}
	\caption{Artificial neuron model in ANNs. }
	\label{Fig:neuron}
\end{figure}
Neurons in ANNs take inputs from their previous layer, and feed the weighted sum of their input, $net_j = \sum_i w_{ij}x_i$, to the activation function.
The transformed signal then forms the output of an artificial neuron, $y_j=f(net_j)$, see Figure~\ref{Fig:neuron}.	
However, Noisy Softplus takes physical quantities of current, and firing rate as input/output, thus an extra step is still needed to map the firing rate to numerical values in ANNs.
According to Equation~(\ref{equ:distr}), the mean of the current feeding into a spiking neuron is equivalent to $net$ of artificial neurons, where
\begin{equation}
\begin{aligned}
& {m_I}_j = \sum_i w_{ij}(\lambda_{i}\tau_{syn})~, \textrm{  then}\\
& net_j= \sum_i w_{ij} x_i~~, \textrm{~~and~~}
x_i = \lambda_{i}\tau_{syn}~.
\end{aligned}
\label{equ:mi_input}
\end{equation}
The noise level of Noisy Softplus, $\sigma^2$, is the variance of the current, which also can be seen as a weighted sum of the same input $x$ but with different weights:
\begin{equation}
\begin{aligned}
& {s_I^2}_j=\sum_i(\frac{1}{2} w_{ij}^2) (\lambda_{i}\tau_{syn})~, \textrm{  then}\\
& \sigma^2_j= \sum_i (\frac{1}{2} w_{ij}^2) x_i~~.
\end{aligned}
\label{equ:si_input}
\end{equation}

\begin{figure}[bt!]
	\centering
	\includegraphics[width=0.98\textwidth]{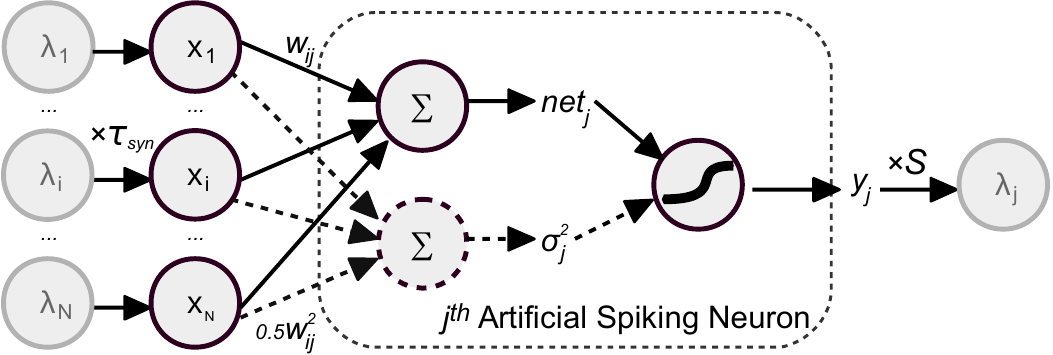}
	\caption{Artificial spiking neuron takes scaled firing rates as input, then transforms weighted sum in some activation unit to its output which can be scaled-up to the firing rate of an output spike train.}
	\label{Fig:sneuron}
\end{figure}

Noisy Softplus transforms the noisy current with parameters of $(net_j, \sigma_j)$ to the equivalent ANN output $y_j$ , where it can be scaled up by the factor $S$ to the firing rate of SNNs.
Note that the calculation of noise level is not necessary for activation functions other than Noisy Softplus, for example, it can be set to a constant for Softplus or 0 for ReLU.
We name the neuron model `artificial spiking neurons' which takes firing rates of spike trains as input and output. 
The entire artificial spiking neuron model is then generalised to any ReLU/Softplus-like activation functions, See Figure~\ref{Fig:sneuron}.

\subsection{Layered-up Network}
\label{subsec:ns_train}
\begin{figure}[tbh!]
	\centering
	\includegraphics[width=0.8\textwidth]{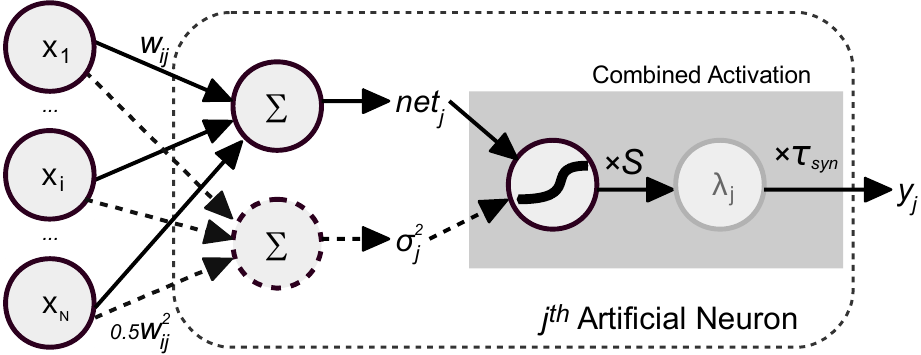}
	\caption{Transforming artificial spiking neurons to artificial neurons for SNN modelling. The combined activation links the firing activity of a spiking neuron to the numerical value of ANNs.}
	\label{Fig:tneuron}
\end{figure}

Referred to Figure~\ref{Fig:sneuron}, if we move the left end process of $\times \tau_{syn}$ to the right end after $\lambda_j$, Figure~\ref{Fig:sneuron} forms the same model structure as artificial neurons shown in Figure~\ref{Fig:neuron}: neurons take $x$ as input and outputs $y$, and this conversion is illustrated in Figure~\ref{Fig:tneuron}.
The process within such an artificial neuron is divided into weighted summation and activation, which also applies to SNN modelling by combining the scaling factor $S$ and the synaptic time constant $\tau_{syn}$ to activation functions.
Thus the combined activation function for modelling SNNs should be:
\begin{equation}
y = f(x) \times S \times \tau_{syn}~~,
\label{equ:full_act}
\end{equation}
and its derivative function which is used when back propagates is:
\begin{equation}
\frac{\partial y}{\partial x} = f'(x) \times S \times \tau_{syn}~~.
\end{equation}

Thus, using this method of ANN-trained SNNs, the activation functions are of lower complexity than the Siegert formula, and their corresponding derivative functions can be directly used for back propagation.
Furthermore, the method enables ReLU-like activation functions for SNN training, thus improving the recognition accuracy while keeping a relative lower firing rate compared to sigmoid neurons. 
Most significantly, the ANN-trained weights are ready for use in SNNs without any transformation, and the output firing rate of a spiking neuron can be estimated in the ANN simulation.

\subsection{Fine Tuning}
There are two aspects to the fine tuning which makes the ANN closer to SNNs:
Firstly, using Noisy Softplus activation functions in a whole trained network operates every single neuron running in a similar noise level as in SNNs, thus the weights trained by other activation functions will be tuned to fit closer to SNNs.
Secondly, the output firing rate of any LIF neuron is greater than zero as long as noise exists in their synaptic input.
Thus adding up a small offset on the labels directs the model to approximate to practical SNNs. 

The labels of data are always converted to binary values for ANN training.
This enlarges the disparities between the correct recognition label and the rest to train the network for better classification capability.
Consequently, we train the network as stated in Section~\ref{subsec:ns_train} with any activation function and then fine-tune it with Noisy Softplus to take account of both accuracy and practical network activities of SNNs.
However, we add a small number, for example 0.01, to all the binary values of the data labels.
Doing so helps the training to loosen the strict objective function to predict exact labels with binary values.
Instead, it allows a small offset to the objective.
An alternative method is to use Softmax function at the top layer, which aims to map real vectors to the range of $(0,1)$ that add up to 1. 
However, without a limit on the input of Softmax, it will be easy to reach or even exceed the highest firing rate of a spiking neuron.
The result of fine tuning on a Convnet will be demonstrated in Section~\ref{subsec:result_compare}.

\section{Results}
\label{sec:iconipResult}
A convolutional network model was trained on MNIST,
a popular database in neuromorphic vision, using the ANN-trained SNN method stated above.
The architecture contains $28\times28$ input units, followed by two convolutional layers 6c5-2s-12c5-2s, and 10 output neurons fully connected to the last pooling layer to represent the classified digit.

The training only employed Noisy Softplus units that all the convolution, average sampling, and the fully-connected neurons use Noisy Softplus function with no bias.
The parameters of the activation function were calibrated as, $(k=0.30, S=201)$,  for LIF neurons~(see~Tablel~\ref{tbl:pynnConfig}) of $\tau_{syn}=5$~ms.
The input images were scaled by 100~Hz to present the firing rates of input spikes.
The weights were updated using a decaying learning rate, 50 images per batch and 20 epochs.
The ANN-trained weights were then directly applied in the corresponding convolutional SNN without any conversion for recognition tasks.

\subsection{Neural Activity}
To validate how well the Noisy Softplus activation fits to the response firing rate of LIF neurons in a real application, we simulated the model on NEST using the Poisson MNIST dataset~\cite{liu2016bench} and the neurons of a convolutional map were observed.

\begin{figure}[tbh!]
	\centering
	\begin{subfigure}[t]{0.8\textwidth}
		\includegraphics[width=\textwidth]{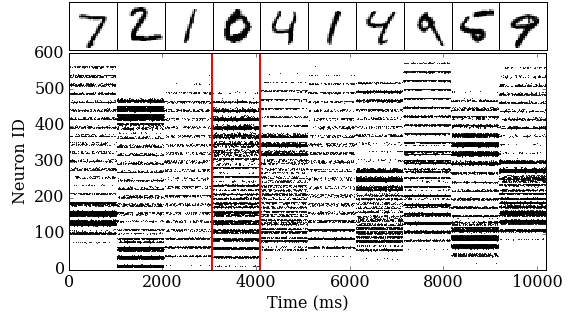}
		\caption{10 input digits presented in Poisson spike trains in raster plot}
		\label{Fig:61}
	\end{subfigure}\\
	\begin{subfigure}[t]{0.3\textwidth}
		\includegraphics[width=\textwidth]{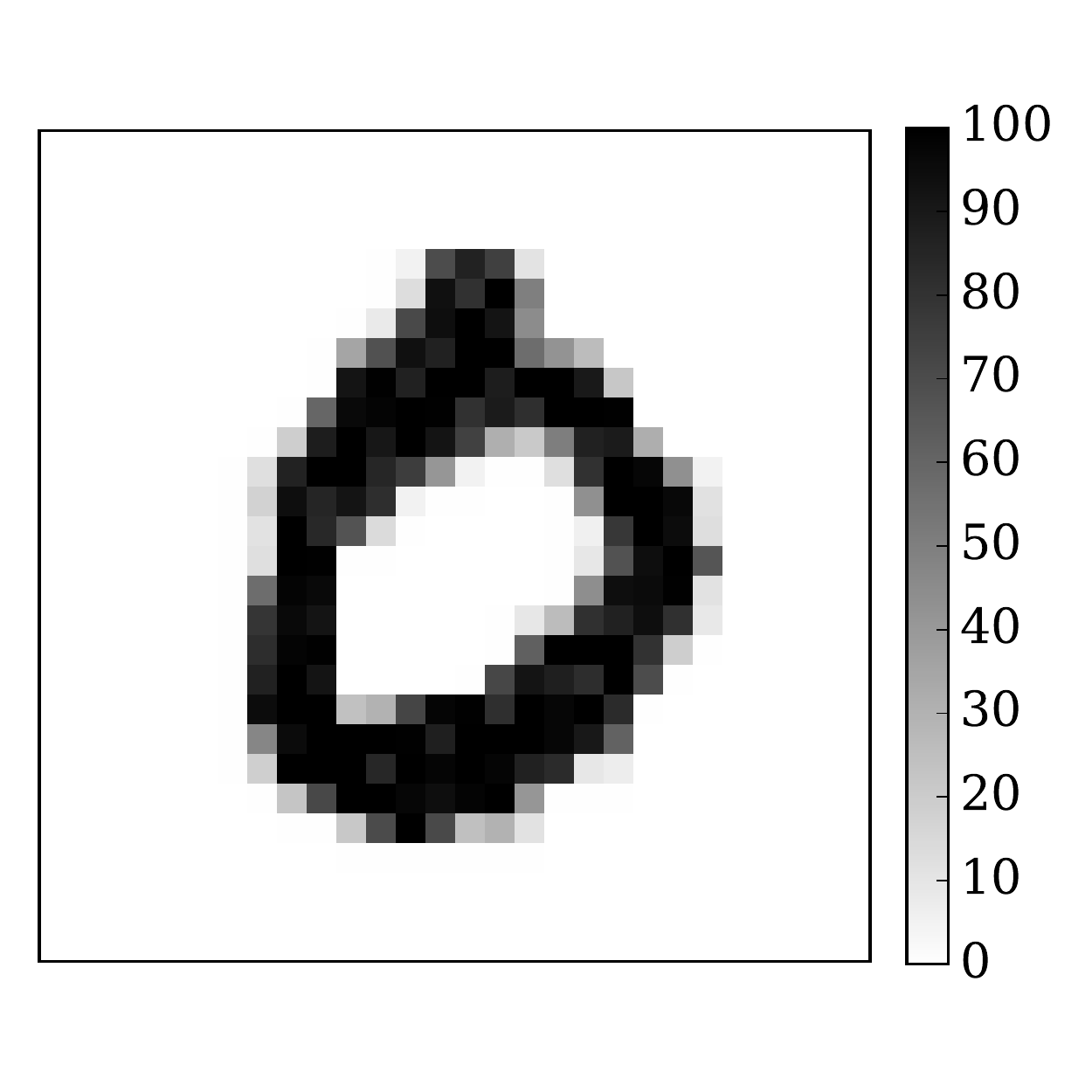}
		\caption{Pixel firing rates}
		\label{Fig:62}
	\end{subfigure}
	\begin{subfigure}[t]{0.3\textwidth}
		\includegraphics[width=\textwidth]{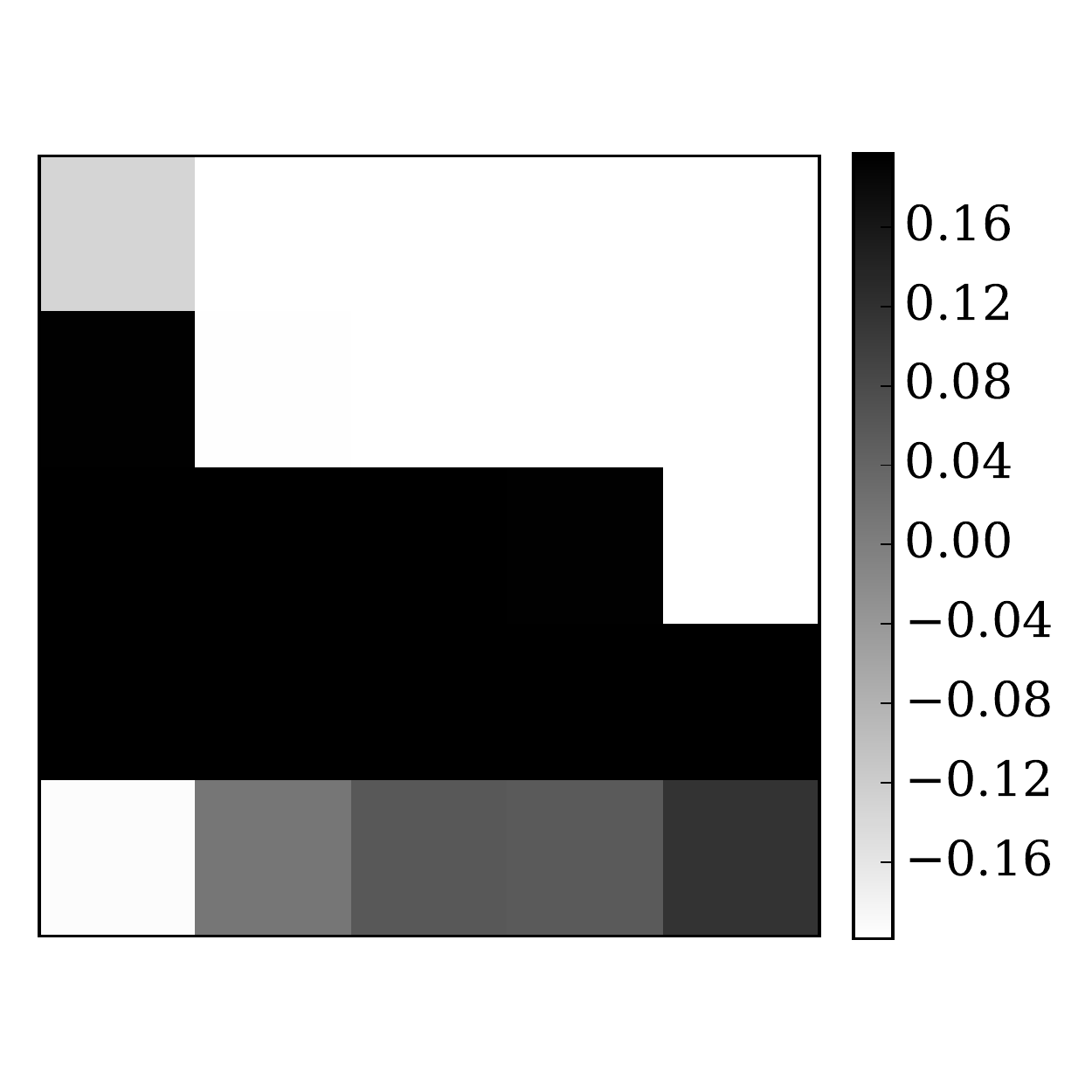}
		\caption{5x5 kernel}
		\label{Fig:63}
	\end{subfigure}
	\begin{subfigure}[t]{0.3\textwidth}
		\includegraphics[width=\textwidth]{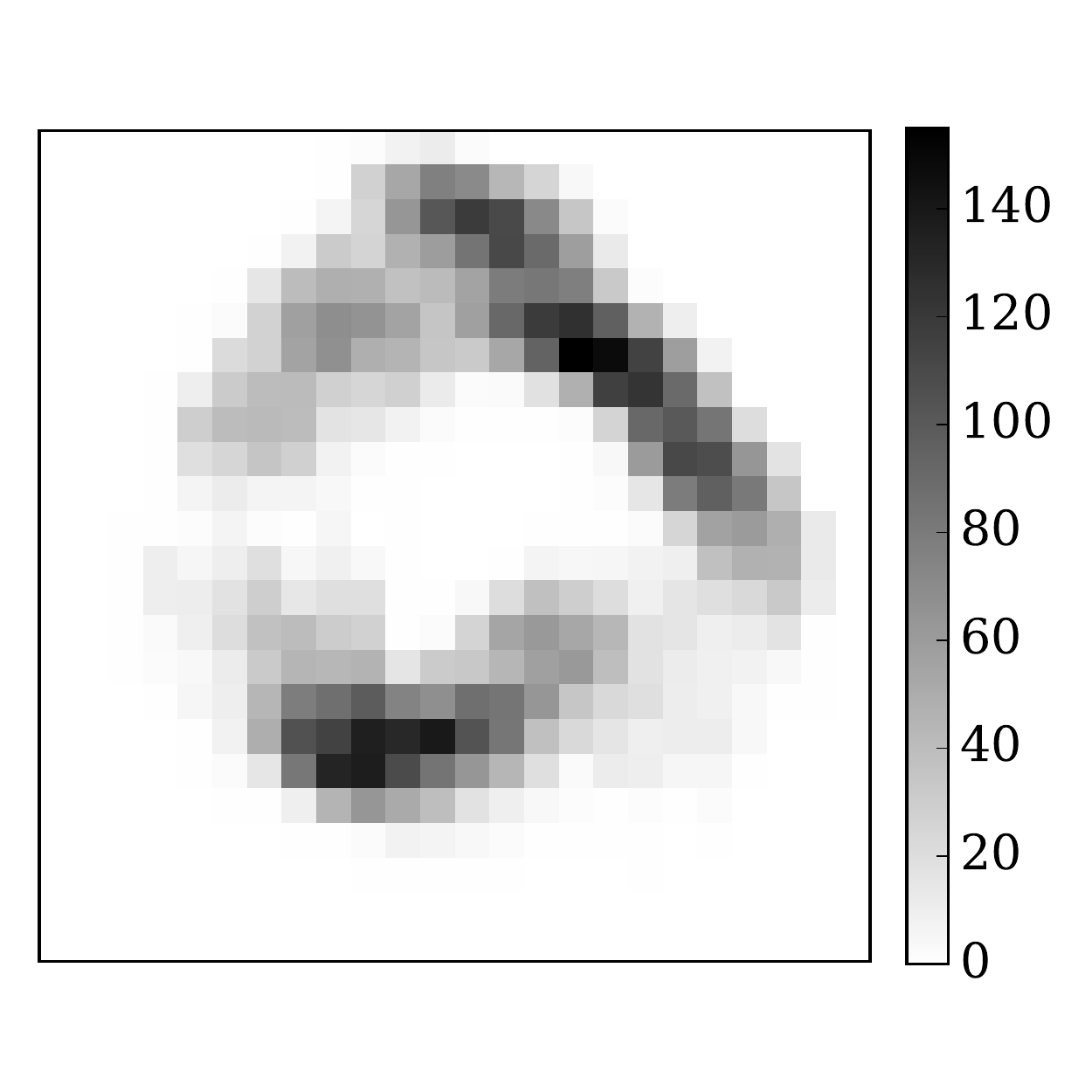}
		\caption{Output firing rates}
		\label{Fig:64}
	\end{subfigure}
	\caption{Images presented in spike trains convolve with a weight kernel. (a) The $28\times28$ Poisson spike trains in raster plot, representing 10 digits in MNIST. (b) The firing rate of all the 784 neurons of the fourth image, digit `0', is plotted as a 2D image.
		(c) One out of six of the trained kernels ($5\times5$ size) in the first convolutional layer.
		(d) The spike trains plotted as firing rate of the neurons in the convolved 2D map.}
	\label{fig:cnn}
\end{figure}

Figure~\ref{fig:cnn} shows a small test of ten MNIST digits presented in Poisson spike trains for 1~s each.
A trained $5\times5$ kernel (Figure~\ref{fig:cnn}(c)) was convolved with these input digits, and the fourth digit `0' (Figure~\ref{fig:cnn}(b)) and its convolved output of the feature map was shown in Figure~\ref{fig:cnn}(d) as firing rate.
The output firing rate was recorded during a real-time SNN simulation on NEST, and compared to the modelled activations of Equation~(\ref{equ:full_act}) in ANNs.

\begin{figure}[tbh!]
	\centering
	\begin{subfigure}[t]{0.6\textwidth}
		\includegraphics[width=\textwidth]{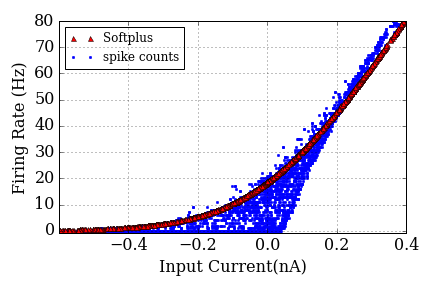}
	\end{subfigure}\\
	\begin{subfigure}[t]{0.6\textwidth}
		\includegraphics[width=\textwidth]{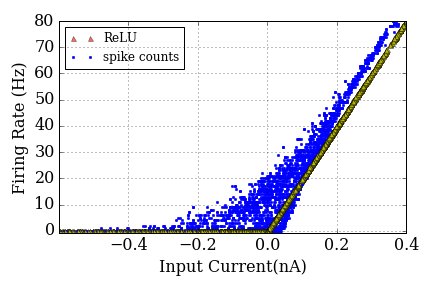}
	\end{subfigure}\\
	\begin{subfigure}[t]{0.6\textwidth}
		\includegraphics[width=\textwidth]{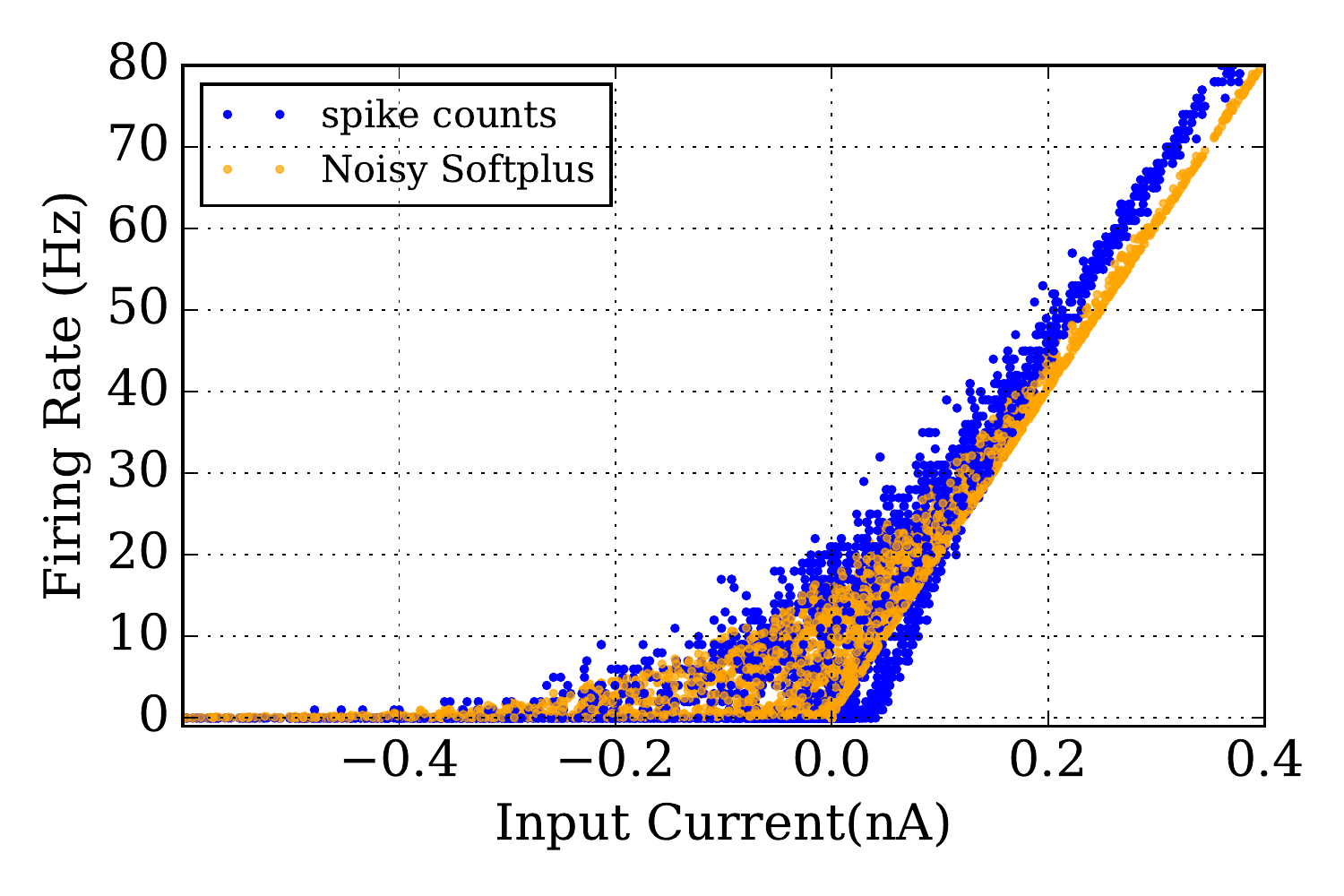}
	\end{subfigure}
	\caption{
		Noisy Softplus fits to the neural response firing rate in an SNN simulation.
		The recorded firing rate of the same kernel convolved with 10 images shown in Figure~\ref{fig:cnn} in SNN simulation, comparing to the prediction of activations of Softplus, ReLU, and Noisy Softplus.}
	\label{fig:af_compare}
\end{figure}

The input $x$ of the network was calculated as Equation~(\ref{equ:mi_input}): $x_i=\lambda_i\tau_{syn}$, and so as the weighted sum of the synaptic current (see Equation~(\ref{equ:mi_input})), $net_j$ and its variance (see Equation~(\ref{equ:si_input})), $\sigma^2_j$.
With three combined activation functions as Equation~(\ref{equ:full_act}):
\begin{equation}
\begin{aligned}
&\textrm{(1) Noisy Softplus:~~}  y_j=k \sigma_j \log [1 + \exp(\frac{net_j}{k \sigma_j})] \times S \times \tau_{syn}~~,  \\
&\textrm{(2) ReLU:~~ } y_j=max(0, net_j) \times S \times \tau_{syn}~~, \\
&\textrm{(3) Softplus:~~ } y_j=k \sigma \log [1 + \exp(\frac{net_j}{k \sigma})] \times S \times \tau_{syn}~~, ~~~\sigma=0.45,  
\end{aligned}
\end{equation}	
we compare the output to the recorded SNN simulations.
ReLU assumes a non-noise current, and Softplus takes a static noise level thus $\sigma_j$ is not used for either of them, meanwhile Noisy Softplus adapts to noise automatically with $\sigma_j$.
The experiment took the sequence of 10 digits shown in Figure~\ref{fig:cnn}(a) to the same kernel and the estimated spike counts using Noisy Softplus fit to the real recorded firing rate much more accurately than ReLU and Softplus,  see~\ref{fig:af_compare}.
The Euclidean distance, $\sqrt{\sum_{j}(y_j/\tau_{syn} - \lambda_j)}$, between the spike counts and the predicted firing rates by Noisy Softplus, ReLU and Softplus was 184.57, 361.64 and 1102.76 respectively.
We manually selected a static noise level of 0.45 for Softplus, whose estimated firing rates located roughly on the top slope of the real response activity.
This resulted in longer Euclidean distance than using ReLU, since most of the input noisy currents were of relatively low noise level in this experiment.
Hence, the firing rate driven by lower noise level is closer to ReLU curve than Softplus.

\begin{figure}[tbp!]
	\centering
	\includegraphics[width=0.5\textwidth]{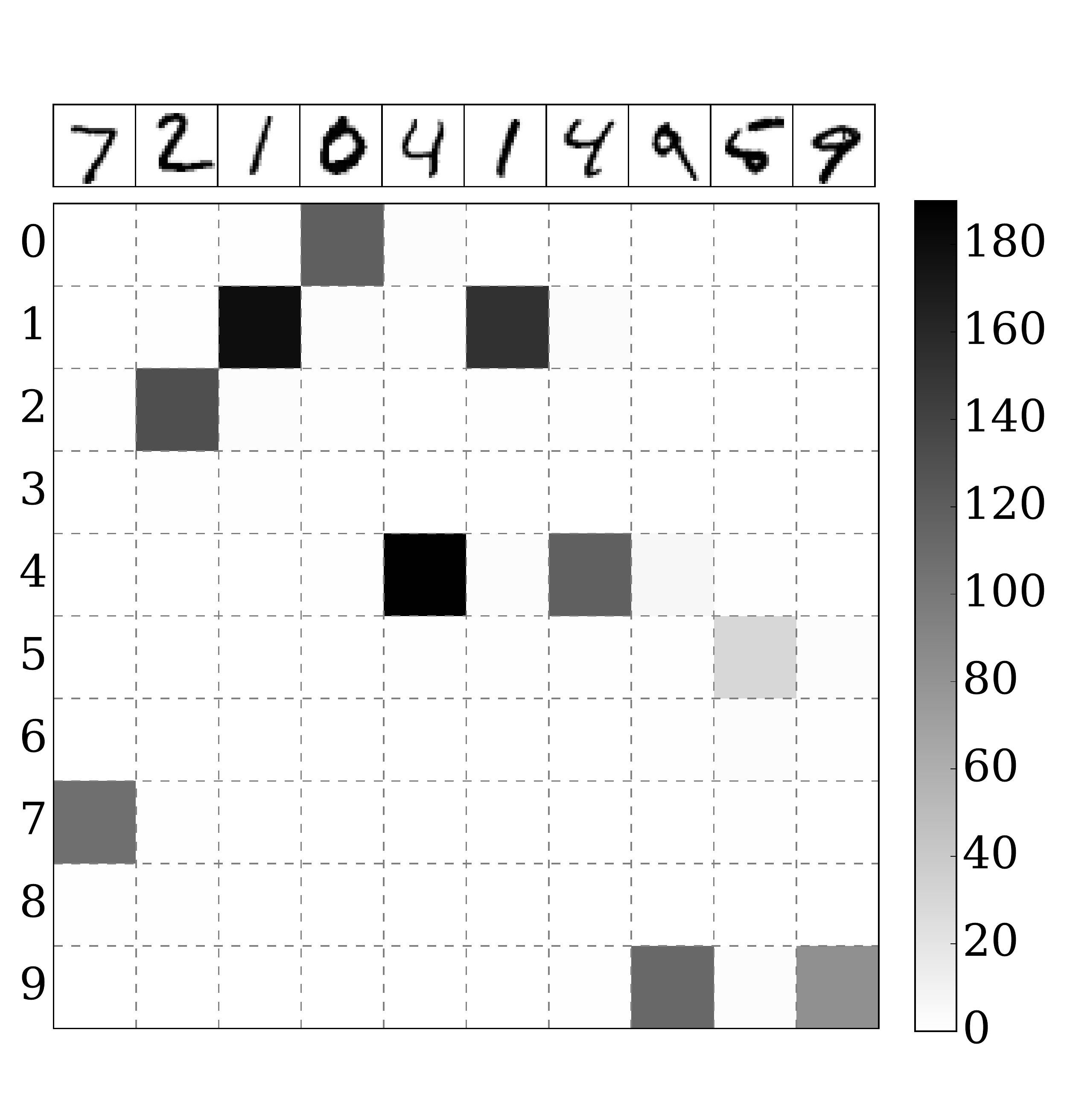}
	\caption{Output firing rates for recognising 10 hand written digits.}
	\label{Fig:out}
\end{figure}

Figure~\ref{Fig:out} demonstrates the output firing rates of the 10 recognition neurons when tested with the digit sequence.
The SNN successfully classified the digits where the correct label neuron fired the most.
We trained the network with binary labels on the output layer, thus the expected firing rate of correct classification was $1/\tau_{syn}=200$~Hz according to Equation~(\ref{Fig:tneuron}).
The firing rates of the recognition test fell to the valid range around 0 to 200~Hz.
This shows another advantage of the proposed ANN-trained method that we can constrain the expected firing rate of the top layer, thus preventing SNN from exceeding its maximum firing rate, for example 1000~Hz when time resolution of SNN simulation set to 1~ms.

\subsection{Recognition Performance}
\label{subsec:result_compare}
\begin{figure}[tbp!]
	\centering
	\includegraphics[width=0.7\textwidth]{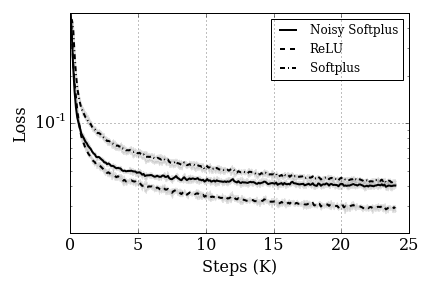}
	\caption{Comparisons of Loss during training using Noisy Softplus, ReLU and Softplus activation functions. Bold lines show the average of three training trials, and the grey colour illustrates the range between the minimum and the maximum values of the trials.  }
	\label{Fig:loss_ns}
\end{figure}

Here we focus on the recognition performance of the proposed ANN-trained SNN method.
Before looking into the recognition results, it is significant to see the learning capability of the proposed activation function, Noisy Softplus.
We compared the training using ReLU, Softplus, and Noisy Softplus by their loss during training averaged over 3 trials, see Figure~\ref{Fig:loss_ns}.
ReLU learned fastest with the lowest loss, thanks to its steepest derivative.
In comparison, Softplus accumulated spontaneous firing rates layer by layer and its derivative may experience vanishing gradients during back propagation, which result in a more difficult training.
Noisy Softplus performance lay between these two in terms of loss and learning speed.
However, the loss stabilised fastest, which means a possible shorter training time.

The recognition test took the whole testing dataset of MNIST which contains $10,000$ images.
At first, all trained models were tested on the same artificial neurons as used for training in ANNs, and these experiments were called `DNN' test since the network had a deep structure (5 layers).
Subsequently, the trained weights were directly applied to SNN without any transformation, and these `SNN' experiments tested their recognition performance on the NEST simulator.
The LIF neurons had the same parameters as in training.
The input images were converted to Poisson spike trains and presented for 1~s each.
The output neuron which fired the most indicated the classification of an input image.
Moreover, a `Fine tuning' test took the trained model for fine tuning, and the tuned weights were tested on the same SNN environment.
The tuning only ran for one epoch, 5\% cost of the ANN training (20~epochs), using Noisy Softplus neurons with labels shifted for $+0.01$.

\begin{figure}[hbt!]
	\centering
	\includegraphics[width=0.7\textwidth]{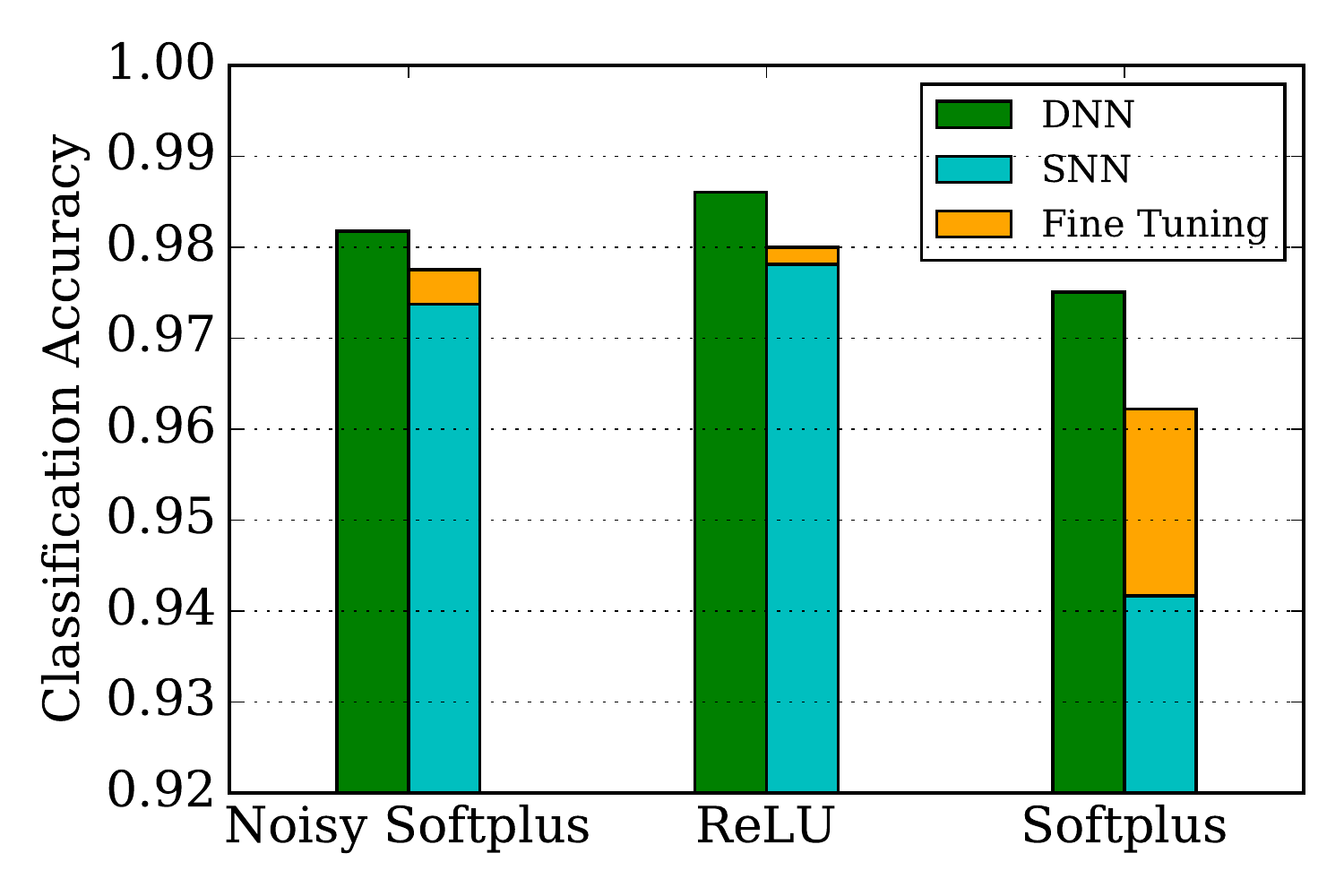}
	\caption{Classification accuracy compared among trained weights of Noisy Softplus, ReLU, Softplus on DNN, SNN and fine-tuned SNN.}
	\label{Fig:result_bar}
\end{figure}

The classification errors for the tests are investigated in Table~\ref{tbl:ns_result} and the averaged classification accuracy is shown in Figure~\ref{Fig:result_bar}.
From DNN to SNN, the classification accuracy declines by 0.80\%, 0.79\% and 3.12\% on average for Noisy softplus, ReLU and Softplus
The accuracy loss was caused by the mismatch between the activations and the practical response firing rates, see example in Figure~\ref{fig:af_compare}, and the strict binary labels for Noisy Softplus and Softplus activations.
Fortunately, the problem is alleviated by fine tuning which increased the classification accuracy by 0.38\%, 0.19\% and 2.06\%, and resulted in the total loss of 0.43\%, 0.61\%, and 1.06\% respectively.
The improvement of ReLU is not as great as the others, because there is no problem of strict labels during training.
Softplus benefits the most from fine tuning, since not only the huge mismatch of response firing rate is greatly corrected, but also the offset on the labels helps the network to fit SNNs. 

\begin{table}[tbh] 
	\caption{Comparisons of classification accuracy (in \%) of ANN-trained convolutional neural models on original DNN, NEST simulated SNN, and SNN with fine-tuned (FT) model.}
	\begin{center}
		\bgroup
		\def\arraystretch{1.5}
		\begin{tabular} {r |c  c c |c c c |c c c}
			\hline
			Trial No.
			&\multicolumn{3}{c|}{1} 
			&\multicolumn{3}{c|}{2}
			&\multicolumn{3}{c}{3}\\
			\hline
			Model
			& DNN & SNN &FT
			& DNN & SNN &FT
			& DNN & SNN &FT\\
			\hline
			\textbf{Noisy Sofplus}
			& 1.91 & 2.76 &2.45
			& 1.79 & 2.56 &2.19
			& 1.76 & 2.55 &2.10\\
			\textbf{ReLU}
			& 1.36 & 2.03 &1.88
			& 1.46 & 2.28 &2.00
			& 1.36 & 2.25 &2.12\\
			\textbf{Sofplus}
			& 2.30 & 5.66 &3.91
			& 2.75 & 5.22 &3.55
			& 2.42 & 6.62 &3.87\\
			\hline
		\end{tabular}
		\egroup
		\label{tbl:ns_result}
	\end{center}
\end{table}

The most efficient training in terms of both classification accuracy and algorithm complexity, takes ReLU for ANN training and Noisy Softplus for fine tuning.
Softplus does not exhibit better classification capability and more importantly the manual selected static noise level hugely influences the mismatch between the predicted firing rates and the real data.
Although Noisy Softplus shows the least classification drop from ANNs to SNNs, the training performance is still worse than ReLU.

The best classification accuracy achieved by SNN was 98.85\%, a 0.20\% drop from ANN test (99.05\%), which was trained with ReLU and fine-tuned by Noisy Softplus.
The network structure was the same with the state-of-the-art model which reported the best classification accuracy of 99.1\%~\cite{diehl2015fast} in ANN-trained SNNs: 12c5-2s-64c5-2s-10fc.
Their nearly loss-less conversion from ANNs to SNNs was achieved by using IF neurons, while our network performs the best among SNNs consisted of LIF neurons to our knowledge.

\begin{figure}[htb!]
	\centering
	\begin{subfigure}[t]{0.49\textwidth}
		\includegraphics[width=\textwidth]{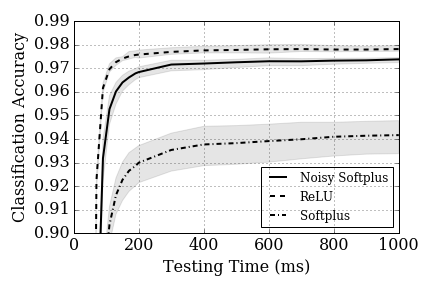}
		\caption{Before fine tuning}
	\end{subfigure}
	\begin{subfigure}[t]{0.49\textwidth}
		\includegraphics[width=\textwidth]{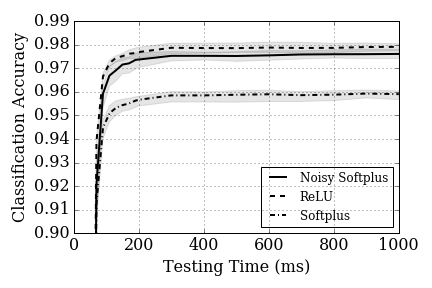}
		\caption{After fine tuning.}
	\end{subfigure}
	
	\caption{The classification accuracy of 3 trials (averaged in bold lines, grey shading shows the range between minimum to maximum) over short response times, with (a) trained weights before fine tuning, and (b) after fine tuning.}
	\label{fig:ca_time}	
\end{figure}

As it is a major concern in neuromorphic vision, the recognition performance over short response times is also estimated in Figure~\ref{fig:ca_time}.
After fine tuning, Softplus significantly reduced the mismatch since the randomness among the three trials shrinks to a range similar to other experiments.
More obviously, fine tuning improved its classification accuracy and the response latency.
Notice that all of the networks trained by three different activation functions showed a very similar stabilisation curve against time, which means they all reached an accuracy close to their best by only taking 300~ms of test.

\subsection{Power Consumption}
Noisy Softplus can easily be used for energy cost estimation for SNNs.
For a single neuron, the energy consumption of the synaptic events it triggers is:
\begin{equation}
\begin{aligned}
E_{j} &= \lambda_j N_j T E_{syn}\\
&= \dfrac{y_j N_j T E_{syn}}{\tau_{syn}}~~,
\end{aligned}
\label{equ:energy}
\end{equation}
where $\lambda_j$ is the output firing rate, $N_j$ is the number of post-synaptic neurons it connects to, $T$ is the testing time, and $E_{syn}$ is the energy cost for a synaptic event of some specific neuromorphic hardware, for example, about 8~nJ on SpiNNaker~\cite{stromatias2013power}.
Thus to estimate the whole network, we can sum up all the synaptic events of all the neurons:
\begin{equation}
\sum_j E_{j} =  \dfrac{T E_{syn}}{\tau_{syn}} \sum_{j}y_j N_j.
\end{equation}
Thus, it may cost SpiNNaker 0.064~W, 192~J running for $3,000$~s with synaptic events of $8\times10^6/s$ to classify $10,000$ images (300~ms each) with an accuracy of 98.02\%.
The best performance reported using the larger network may cost SpiNNaker 0.43~W operating synaptic event rate at $5.34\times10^7/s$, consume 4271.6~J to classify all the images for 1~s each.

\section{Summary}
Most significantly, we proposed the Noisy Softplus activation function which accurately models response firing rate of LIF neurons and overcomes the drawbacks of Siegert units.
%
%
%
%

Moreover, we proposed complete SNN modelling method by using artificial neurons of combined activation;
this method can be generalised to activation units other than Noisy Softplus.
The training of an SNN model is exactly the same as ANN training, and the trained weights can be directly used in SNN without any transformation.
This method is simpler and even more straight-forward than the other ANN offline training methods which requires an extra step of converting ANN-trained weights to SNN's.

In terms of classification/recognition accuracy, the performance of ANN-trained SNNs is nearly equivalent as ANNs, and the performance loss can  be partially solved by fine tuning.
The best classification accuracy of 98.85\% using LIF neurons in PyNN simulation outperforms state-of-the-art SNN models of LIF neurons and is very close to the result using IF neurons~\cite{diehl2015fast}.
 
\section*{Acknowledgments}
The research leading to these results has received funding from the European Research Council 
(FP/2007-2013) / ERC Grant Agreement n. 320689 and from the EU Flagship Human Brain Project (FP7-604102). 
Yunhua Chen received funding from the Natural Science Foundation of Guangdong Province, China (No: 2016A030313713) and also from  the Natural Science Foundation of Guangdong Province, China (No: 2014A030310169).
Qian Liu personally thanks to the funding from the National Natural Science Foundation of China (61662013,U1501252), the Guangxi Natural Science Foundation~(2014GXNSFDA118036), and The High Level of Innovation Team of Colleges and Universities in Guangxi and Outstanding Scholars Program Funding.
\bibliography{ref}
\bibliographystyle{splncs03}
\end{document}